\documentclass{article}

\usepackage{amsmath,amsfonts,bm}

\def\eqref#1{equation~\ref{#1}}

\def\1{\bm{1}}

\DeclareMathAlphabet{\mathsfit}{\encodingdefault}{\sfdefault}{m}{sl}
\SetMathAlphabet{\mathsfit}{bold}{\encodingdefault}{\sfdefault}{bx}{n}

\def\sI{{\mathbb{I}}}

\DeclareMathOperator*{\argmax}{arg\,max}
\DeclareMathOperator*{\argmin}{arg\,min}

\usepackage{microtype}
\usepackage{graphicx}
\usepackage{subfigure}
\usepackage{booktabs} 
\usepackage{multirow}
\usepackage{adjustbox}
\usepackage{makecell}
\usepackage[most]{tcolorbox}
\usepackage{tabularx}
\usepackage{longtable}
\usepackage{pifont}
\usepackage{caption}
\usepackage{listings}
\usepackage{hyperref}
\usepackage{lmodern}

\usepackage[accepted]{icml2024}

\usepackage{amsmath}
\usepackage{amssymb}
\usepackage{mathtools}
\usepackage{amsthm}
\usepackage{microtype} 
\usepackage[capitalize,noabbrev]{cleveref}

\theoremstyle{plain}

\theoremstyle{definition}

\theoremstyle{remark}

\usepackage[textsize=tiny]{todonotes}

\icmltitlerunning{Toward Adaptive Reasoning in Large Language Models with Thought Rollback}

\begin{document}

\twocolumn[
\icmltitle{Toward Adaptive Reasoning in Large Language Models with Thought Rollback}




\begin{icmlauthorlist}
    \icmlauthor{Sijia Chen}{yyy}
    \icmlauthor{Baochun Li}{yyy}
\end{icmlauthorlist}

\icmlaffiliation{yyy}{Department of Electrical and Computer Engineering, University of Toronto, Toronto, Ontario, Canada}

\icmlcorrespondingauthor{Sijia Chen}{sjia.chen@mail.utoronto.ca}

\icmlkeywords{Large Language Models, Multi-step Reasoning}

\vskip 0.3in
]



\printAffiliationsAndNotice{} 

\begin{abstract}
    Large language models (LLMs) have been routinely used to solve various tasks using step-by-step reasoning. However, the structure of intermediate reasoning steps, or \emph{thoughts}, is rigid and unidirectional, such as chains, trees, or acyclic-directed graphs. Consequently, the resulting inflexible and forward-only reasoning may not address challenging tasks and fail when the LLM frequently gives false responses, i.e., ``hallucinations''. This paper proposes a new reasoning framework, called \emph{Thought Rollback} (TR), allowing LLMs to adaptively build thought structure while maintaining effective reasoning toward problem-solving under ``hallucinations''. The core mechanism of TR is \emph{rolling back thoughts}, which allows LLMs to perform error analysis on thoughts, and thus roll back to any previously mistaken thought for revision. Subsequently, by including such trial-and-error in the prompt to guide the LLM, each rollback leads to one more reliable reasoning path. Therefore, starting with a simple prompt without human annotations, LLM with TR adaptively and gradually explores thoughts for a correct solution. Comprehensive experiments on mathematical problems and multi-task reasoning demonstrate the state-of-the-art performance of TR in terms of problem-solving rate and interaction cost. For instance, the solving rate of GPT-4 with TR outperforms the current best by $9\%$ on the \texttt{MATH} dataset. The source code is available under the folder \textit{examples/ThoughtRollback} of \url{https://github.com/iQua/llmpebase}.
\end{abstract}

\section{Introduction}
\label{sec:intro}

Large Language Models, initially designed for text generation with autoregression, are widely recognized to excel in a diverse array of natural language processing (NLP) tasks. Yet, at a particular model scale, their reasoning abilities, particularly in scaled-up versions like GPT-4 \cite{gpt4-arxiv23} and Llama 2 \cite{llama2-arxiv23}, heavily depend on prompt engineering. With well-crafted prompts --- even just a simple \emph{Let's think step by step} \cite{llmszeroshot-nips22} --- LLMs are able to perform step-by-step reasoning and achieved noteworthy success in mathematical, symbolic, and common sense tasks. With reasoning, LLMs are capable of producing coherent language sequences, called \emph{thoughts}, which serve as intermediate reasoning steps toward solving the problem at hand. Extended from simple chain reasoning \cite{cot-nips22} with linear left-to-right thoughts, more complex reasoning became feasible in recent works by establishing thought structures that resembled trees \cite{tot-nips23} and graphs \cite{graphthought-arxiv23,cr-arxiv23,rog-iclr24}.

\begin{figure*}[h]
    \centering
    \includegraphics[width=\textwidth]{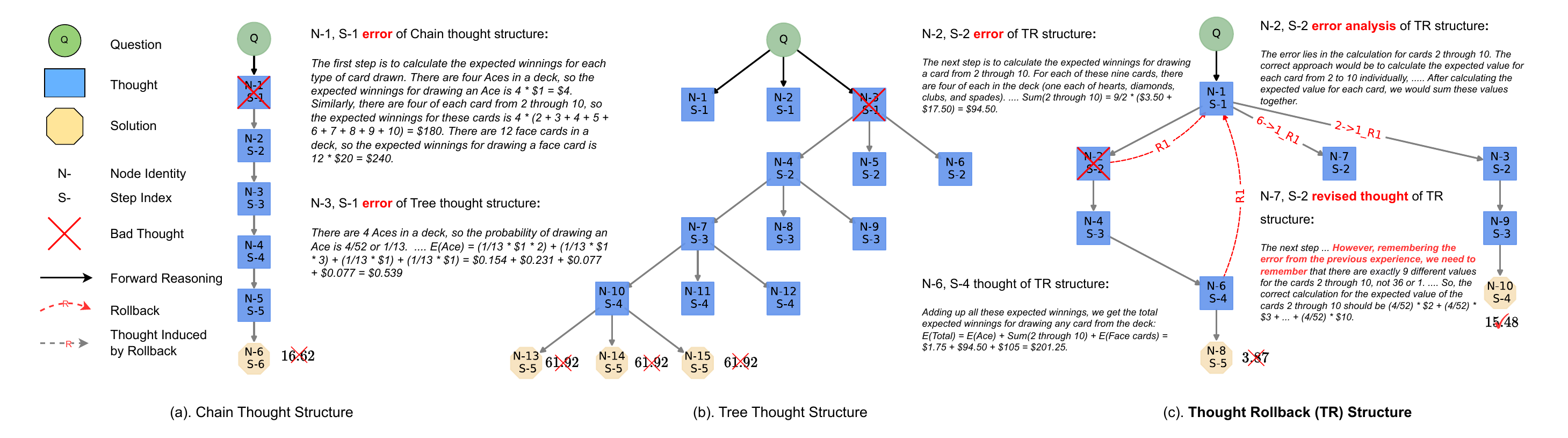}
    \caption{Schematic illustrating thought structures for problem solving with GPT-4. The chain, tree, and our thought rollback (TR) structures are plotted based on the NetworkX lib \cite{netx-08}. The question from the \texttt{MATH} dataset \cite{MATH-arxiv21} is: \textit{I draw a card from a standard 52-card deck. If I draw an Ace, I win 1 dollar. If I draw a 2 through 10, I win a number of dollars equal to the value of the card. If I draw a face card (Jack, Queen, or King), I win 20 dollars. If I draw a `clubsuit', my winnings are doubled, and if I draw a `spadesuit', my winnings are tripled. ... What would be a fair price to pay to play the game?} In (c), we present a partial thought structure built by GPT-4 with TR and place the full version of the TR structure in Figure \ref{fig:introcomplete} of the Appendix.}
    \label{fig:intro}
    \vspace{-0.4cm}
\end{figure*}

However, existing thought structures are unidirectional and thus allow a forward-only reasoning process, meaning that thoughts are generated sequentially from the start to the end. The efficacy of this reasoning process hinges on a redundant and, consequently, inefficient thought structure, requiring thorough explorations of each thought before progressing to the next. One major drawback of forward-only reasoning is that errors can propagate quickly \cite{tp-iclr24}. Consider the common case where one thought is incorrect or inaccurate: with forward-only reasoning, all thoughts derived from it can be misled. Even with revisions based on step-by-step evaluations \cite{selfverification-emnlp23}, such propagation of errors can introduce further deviations from the correct path of reasoning, since LLMs have been found to confidently provide false information (i.e., ``hallucinations'' or ``laziness'') \cite{cape-icml23}. 

Indeed, humans also provide false information as frequently and randomly as LLMs do during reasoning but can still solve challenging problems eventually. This is attributed to adaptive reasoning, in which one does not pre-defined a fixed structure for thoughts and does not simply deduce forward but adaptively adjusts the thought structure after evaluating errors during reasoning. Such reasoning enables humans to begin with one simple or wrong thought but frequently introspect during reasoning, that is, to reconsider previous steps and build new reasoning paths from these reflections. In this paper, we argue that this iterative error correction nature of adaptive reasoning is essentially supported by \textbf{rollback} --- jumping to previous steps with a new experience to reconsider the reasoning.

Therefore, we propose a new reasoning framework, Thought Rollback (TR), relying upon the rolling back of thoughts to enable the adaptive reasoning of LLMs for general problem solving. TR embraces a \textit{rollback controller} and a \textit{prompt enhancer} that works seamlessly to enable the LLMs to generate an effective thought structure from one thought derived from a simple input prompt, as shown by Figure~\ref{fig:intro}. 

LLMs with TR start with generating one thought from a simple zero-shot prompt containing only the question description. Subsequently, for each generated thought, the \textit{rollback controller} allows the LLM to analyze the obtained chain of thoughts and thus determines whether to roll back and to which previous thought. Once rollback is triggered, \textit{prompt enhancer} accumulates this error analysis as experience in the prompt. As a result, by avoiding making similar mistakes mentioned by experience, LLM is able to generate a new and more effective reasoning path from the chosen thought. Therefore, ``hallucinations'' that occur in thought or analysis of LLM may not influence reasoning due to the continuous thought revision guaranteed by the iterative rollbacks during reasoning. For example, in Figure~\ref{fig:intro}, different from chain \cite{cot-nips22} and tree \cite{tot-nips23} structures, which assume a fixed and unidirectional structure, reasoning with rollbacks enables LLM to build a thought structure adaptively and revise thoughts to achieve complex but reliable reasoning. Specifically, after reaching the $N$-$2$ $S$-$2$ and $N$-$6$ $S$-$4$, LLM finds an error in $2$-th step $N$-$2$ $S$-$2$ and thus rolls back to $1$-th step $N$-$1$ $S$-$1$ to create two new reasoning paths. The rollback $N$-$6$ $\rightarrow$ $N$-$1$ leads to the revised thought $N$-$7$ $S$-$2$. The rollback $N$-$2$ $\rightarrow$ $N$-$1$ utilizes the error analysis to enhance the prompt and obtains $N$-$3$ $S$-$2$, leading to a correct answer $15.48$ compared to the previous mistaken $3.87$.

We observe four contributions of TR. First, it is a lightweight and generalized automatic prompting framework. TR allows LLMs to perform complex reasoning effectively on various tasks without introducing task-specific human annotations in the prompt or additional human-made designs. Second, the performance of TR is robust to the ``hallucinations'' as LLMs are able to reconsider and revise any existing thoughts adaptively and repeatedly during reasoning. Thus, third, TR is cost-friendly as the thought structure is built progressively to reach a solution instead of relying on bulky search structures \cite{tot-nips23} or question analogies \cite{tp-iclr24}. Finally, our evaluation of TR on mathematical problems and multi-task reasoning demonstrates that TR outperforms some state-of-the-art approaches while maintaining a lower interaction cost. 

\section{Related Work}
By only guiding the reasoning behavior of LLMs, such as GPT-4 \cite{gpt4-arxiv23} and Llama2 \cite{llama2-arxiv23} with the text prompt, prompt engineering \cite{llmsfewshot-iclr20,llmszeroshot-nips22} is parameter efficient and often matches or surpasses fine-tuning performance. Therefore, plenty of work has been proposed to enable LLMs to perform \textbf{multi-step reasoning} containing a series of intermediate steps, each known as the thought presented as a text sequence. Starting from chain-of-thought (CoT) \cite{cot-nips22} prompting, which provides reasoning examples in the prompt to deduce a chain of thoughts, subsequent endeavors, especially SC \cite{self-consistency-iclr22} and Complex CoT \cite{complexity-iclr23} augment the chain reasoning. Recent advances extend the chain structure into structured thoughts. ToT \cite{tot-nips23} and BoT \cite{bot-iclr24} pre-defines the thought structure as a tree, thus supporting exploring multiple thoughts in each step before generating the next. The graph of thoughts (GoT) \cite{graphthought-arxiv23} and cumulative reasoning (CR) \cite{cr-arxiv23} further instantiate thoughts toward a solution as the graph structure. Another line of work focuses on the thought structure that allows the thought ensemble in each step. Thought propagation \cite{tp-iclr24} explores analogous problems and then aggregates their results to update the solution for the given question, leading to a radial thought structure. To the best of our knowledge, none of the existing work supports the cyclic structure to allow LLMs to revise previous thoughts or recreate a new reasoning path from the previous step after being blocked at the current reasoning step. We fill this gap by proposing the rollback of thought, leading to a thought structure of directed cyclic graphs.

Despite these achievements, LLMs often struggle with complex tasks, primarily due to the frequent occurrence of ``hallucinations''—producing false outputs \cite{cape-icml23,tp-iclr24}, and ``laziness''—yielding invalid or no output. Therefore, after noticing that LLMs have \textbf{self-verification} \cite{selfverification-emnlp23,selfrefine-arxiv23} abilities and thus can analyze the answer for further correcting the errors in reasoning \cite{php-arxiv23, PRPrompt-aaai24}. However, the most recent work \cite{selfcorrection-iclr24} argues that LLMs cannot self-correct their reasoning, emphasizing the invalidity of applying simple verification to the reasoning path. Thus, most recent work either builds iterative-based verification (BoT \cite{bot-iclr24}) or focuses on \textbf{step-by-step verification} \cite{np-nips23,letverify-iclr24}. Combining these insights, we aim to exploit LLMs to analyze intermediate thoughts during reasoning to correct these thoughts and adjust reasoning direction adaptively. Continuous verification and revision may eliminate the negative impact of ``hallucinations'' or mistakes on the solutions. 

Another related research stream is \textbf{automatic prompting} \cite{llmszeroshot-nips22,autocot-iclr22}, which automatically constructs effective prompts to facilitate reasoning without human-made and task-specific demonstrations. As LLMs can learn from mistakes to become better reasoners \cite{mistakes-arxiv23,bot-iclr24}, this paper also releases human efforts from the prompt design by boosting the prompt with the error analysis of thoughts. We also show that by accumulating error analysis in the prompt during reasoning, LLMs are able to avoid making similar mistakes and explore correct solutions with interaction cost far less than ToT \cite{tot-nips23} and BoT \cite{bot-iclr24}. 

\section{Preliminary}
 
\subsection{Problem Statement}

Given a pre-trained large language model (LLM) denoted as $f\left(\cdot\right)$, the prompt engineering is to design the prompt $\sI\left(\cdot\right)$ to make the model perform desired reasoning behavior toward addressing the given problem $x$. Specifically, multi-step reasoning contains $T$ intermediate steps $z_{0...T}=\left[z_0, z_1, ..., z_{T}\right]$ to bridge $z_0:=x$ and the answer $z_T:=y$. To get $z_{0...T}$, we focus on step-wise thought generation in which each thought is a coherent language sequence $z_n$, behaving as one intermediate reasoning step, and $z_n$ is generated as $z_n \sim f\left(z_n|\sI\left(z_{0,1...n-1}\right)\right)$. Therefore, as thought is the LLM's output, we can define the bad thought caused by the ``hallucinations'', ``laziness'', or false reasoning of LLMs as $\widehat{z}_n$. 

These generated thoughts $z_{0...T}$ naturally follow a specific structure, such as a chain or tree. These structures are unidirectional and thus only support \textbf{forward-only reasoning}, which proceeds in a linear, sequential manner, meaning that LLMs only generate the subsequent thought $z_{n+1}$ from $z_n$. For instance, any edge $e_{n,m}$ of the structure is limited to $m = \left\{n, n+1\right\}$ while $m>=n$. 

This paper focuses on alleviating the effect of bad thoughts on the solution by making LLM not simply perform forward-only reasoning but achieve \textit{adaptive reasoning}, which allows LLMs to 1) start from a simple prompt $\sI\left(z_0\right)$, 2). self-organize the thought structure adaptively during reasoning, and thus 3). when $\widehat{z}_n$ occurs, LLM can make revisions and create better new reasoning paths till getting the solution. Specifically, not only advancing the reasoning sequentially, any previously generated thought will be reconsidered by continuously rolling back from $n$-th thought to one previous $m$-th thought, where $m \in \left[0, n-1\right]$. 

\subsection{Motivation: Forward-only reasoning fails in bad thoughts}
\label{subsection:motivation1}

Forward-only reasoning may fail as the bad thought $\widehat{z}_n$ is caused by the following three cases of \textbf{error propagation}.

\textbf{Case $\left[\widehat{z}_m, \widehat{z}_{m+1}, ..., \widehat{z}_{n}\right]$}. A bad or illogical thought $\widehat{z}_m$ leads to all subsequent errors, where $\widehat{z}_n \sim f\left(\widehat{z}_n|\sI\left(\widehat{z}_{0,...m...n-1}\right)\right)$ and $m < n$.

\textbf{Case $\left[\widehat{z}_{m}, z_{m+1}..., \widehat{z}_{n}\right]$}. $\widehat{z}_m$ does not lead to direct mistakes but causes a bad thought $\widehat{z}_n$ after many steps. For instance, this appears when $\widehat{z}_m$ behaves as one part of a solution. 

\textbf{Case $\left[z_{m...n-1}, \widehat{z}_{n}\right]$}. A bad thought $\widehat{z}_n$ may arise from one previous correct though $z_m$ because the wrong reasoning direction appears from this step.

The chain reasoning, typically in Chain-of-thought (CoT) prompting, generates a chain of thoughts $z_{0,1...T}$ \cite{cot-nips22} sequentially; thus, when any cases appear, it gets a mistaken answer $y$. Subsequent Complex CoT \cite{complexity-iclr23} provides well-designed examples in $z_0$ to decrease the frequency of bad thoughts. And SC \cite{self-consistency-iclr22} performs majority voting for many generated chains. They are still trapped in the error propagation of chain structure, as shown in Figure \ref{fig:intro} (a).
 
The tree reasoning, such as Tree of Thoughts (ToT) \cite{tot-nips23}, Graph of Thoughts (GoT) \cite{graphthought-arxiv23}, and Thought Propagation (TP) \cite{tp-iclr24}, extends the chain thought structure by generating multiple thoughts in each step. For instance, ToT or GoT contains $P$ $n$-th thought, representing as $\left(z^{\left(1\right)}_{n},...,z^{\left(P\right)}_{n}\right) \sim f\left(z_{n}|\sI\left(z_{0,1...n-1}\right)\right)$. These approaches hope to explore more thoughts to increase the possibility of getting a correct thought. However, error propagation still exists, and any cases that appear in one reasoning path will inevitably cause the failure, as shown in Figure \ref{fig:intro} (b). 

Therefore, we argue that to address the error propagation, the reasoning should continuously reconsider the source of error, that is, to find and fix the previous $m$-th thought after reaching $n$-th thought, where $m < n$. This is equivalent to \textit{human-like reasoning, in which one may not simply deduce forward but look back to check previous thoughts to decide to continue, revise old thoughts, or create a new reasoning path}. Without losing generality, we refer to such reasoning behavior as the \textbf{rollback}. In the context of reasoning, to enable such a rollback $n \rightarrow m$, we allow LLM to adaptively build the edge $e_{n,m}$ with $m < n$ during reasoning, making the thought structure a directed graph with cycles.

\subsection{Motivation: Error analysis induces better thoughts}
\label{subsection:motivation2}

The rollback mechanism is insufficient to support adaptive reasoning in LLMs. Without introducing more information to prompt LLMs after each rollback, LLMs may repeat similar mistakes in the thought generation, which more rollbacks in reasoning cannot solve. Thus, it is essential to enable LLMs to know why rollback is triggered and how to avoid producing the $\widehat{z}_{m+1}$.

Motivated by the effectiveness of enhancing prompts by including an error report \cite{bot-iclr24}, we propose that analysis of $\left[z_1,...,z_n\right]$ could be rolled back to $z_m$ to guide the thought generation. Also, the work \cite{np-nips23} pointed out that LLMs perform more reliable reasoning in CoT when using step-by-step verification. We aim to allow LLMs to perform rollback-by-rollback verification. First, this can produce more analysis to facilitate the subsequent reasoning. Second, invalid or mistake rollbacks can be removed, thus also eliminating the cycles in the thought structure.

\section{Thought Rollback Framework}
\label{sec:tr}

\subsection{Reasoning Overview}
\label{subsec: overview}

\begin{figure}[t]
    \centering
    \includegraphics[width=\columnwidth]{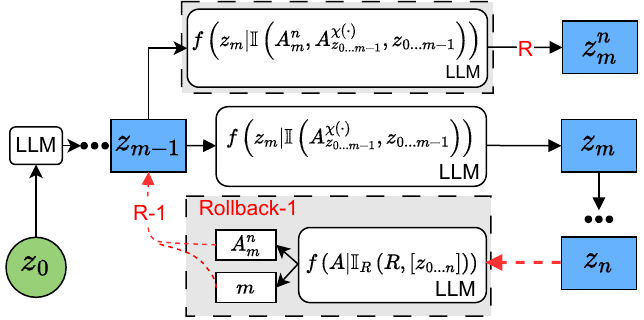}
    \caption{Schematic illustrating the rollback of thought when the LLM with TR reaches the $n$-th reasoning step. We add $A^{\chi\left(\cdot\right)}_{z_{0...m-1}}$ in the reasoning from $z_{m-1}$ to $z_m$ to cover the case that $z_{m-1}$ may also derive from a rollback. We present a clear example from \texttt{SVAMP} in Figure \ref{fig:fig2example} of the Appendix.}
    \label{fig:tr}
    \vspace{-0.4cm}
\end{figure}

In contrast to existing approaches relying on pre-defined unidirectional thought structures, which are limited to forward-only reasoning, Thought Rollback (TR) generates a bidirectional thought structure by adaptively deducing forward and rolling back of thoughts. After reaching a reasoning step $z_n$,  as shown in Figure \ref{fig:tr}, TR allows LLM to roll back to the bad thought $z_m$ after analyzing the existing thought chain $z_{0...n}$. As the error analysis $A_m^n$ is to be accumulated in the prompt, a new and more reliable reasoning path $z_m^n$ is generated from $z_m$. Therefore, iteratively performing this rollback develops the thought structure from a simple thought to a directed graph with cycles. Such adaptive reasoning is summarized into three stages.

\textbf{Initialization}. Generate thought $z_{1} \sim f\left(z_{1}|\sI\left(z_0\right)\right)$. 

\textbf{Rollback of thoughts}. For each generated thought $z_n$, \textit{rollback controller} exploits LLM to determine a rollback to one thought $z_m \in z_{0...n}$.
\begin{itemize}
    \item Once the rollback to $m \in \left[0, 1, ..., n\right]$ is triggered, the reasoning of LLM rolls back to the thought $z_{m-1}$ and \textit{prompt enhancer} is used to enhance the prompt. Subsequently, the reasoning continues by creating a new $m$-th thought $z^{n}_{m}$ and generating $z_{n+1}$, where $z^{n}_{m}$ means a $m$-th thought deduced from a rollback from $n$.
    \item When no rollback is required, generate $z_{n+1}$.
\end{itemize}

\textbf{Early stopping}. Stop reasoning when TR yields $K$ number of solutions obtained. Otherwise, continue the Rollback of thoughts.

\textbf{Solution ensemble}. Perform weighted majority voting on $K$ solutions. 

\subsection{Rolling Back Thought with Reasoning Analysis}
\label{subsec: rollback}
During reasoning, \textit{rollback controller} enables an adaptive rollback by exploiting LLM to determine the rollback $n\rightarrow m$. 
However, making LLM know the concept of rollback may introduce unnecessary complexity. Thus, TR supports the rollback mechanism by performing error analysis on thoughts. Specifically, with a task-agnostic prompt $\sI_R\left(R, \left[z_{0...n}\right]\right)$, where $R$ is a common error analysis instruction, LLM is guided to analyze a thought chain $\left[z_{0...n}\right]$, leading to the error analysis $A^n_m \sim f\left(A|\sI_R\left(R, \left[z_{0...n}\right]\right)\right)$. Eventually, LLM is able to identify the indexes $\widehat{M}$ of bad thoughts $\widehat{z}_{m \in \widehat{M}}$. 

To get which thought to roll back to from $z_n$, TR follows the rule to roll back to the one step before the first bad thought. There are two reasons for this. First, generating the next thoughts from a bad thought is unreasonable. Second, we aim not to remove the bad thought but to generate a new reasoning path. Thus, we choose $z_{m-1}$ as the rollback destination, where $m=\argmin \widehat{M}$. Besides, one thought will not be selected as the rollback destination more than $U$ times to avoid all subsequent thoughts rolling back to the same previous thought. Thus, when the number of rollbacks to a thought reaches $U$, the next earliest one of $m=\argmin \widehat{M} \setminus \left\{\argmin \widehat{M}\right\}$ will be selected. 

Therefore, when $\widehat{M}$ is not empty, TR generates the next thought $z^n_{m}$ from $m-1$, meaning that a new reasoning path $\left[z_0...z_{m-1}, z^n_m\right]$ derived from the rollback $n \rightarrow m$ is created for the thought structure. As the existing $z_{0...n}$ remains unchanged, the reasoning continues by generating $z_{0...n+1}$. For ease of description, we define $n \rightarrow m$ as the \emph{outgoing rollback} for $z_{0...n+1}$ and the \emph{incoming rollback} for the new reasoning path $\left[z_0...z_{m-1}, z^n_m\right]$.

\subsection{Enhancing the Prompt with Errors as Experience}
\label{subsec: enhancing}

Through iterative rollback from the $n$-th to the $m-1$-th step, TR gains the opportunity to address the three scenarios outlined in subsection \ref{subsection:motivation1}. However, as discussed in \ref{subsection:motivation2}, regenerating a next thought $z^n_m$ based on the same prompt may repeat existing mistakes in the new reasoning path. Especially considering that TR is built upon the prompt containing no human annotations, the thought regeneration after the rollback is equivalent to randomly exploring $z^n_m$ as in unidirectional structures.  

Therefore, TR embraces \textit{prompt enhancer} to also roll back the error analysis $A^n_m$ to the $m-1$-th thought. Unlike BoT \cite{bot-iclr24} with outcome analysis, which utilizes error feedback on the final result, TR performs process analysis, i.e., rollback-by-rollback verification, to get error reports on intermediate thoughts, guiding the subsequent thought generation. With error analysis, each rollback is regarded as a trial on generating subsequent thoughts for $z_{m-1}$ because the analysis contains a trial experience: \textit{what mistakes may appear in the following steps of $z_{m-1}$}. By including $A^n_m$ as an experience in the prompt, LLMs can avoid making existing bad thoughts after learning from mistakes. Eventually, each rollback $n \rightarrow m$ creates the error experience $A^n_m$. 

\textbf{Experience accumulation}. The thoughts $\mathbf{Z}^{\chi\left(\cdot\right)}_{z_{0...q-1}} = \left\{z^i_j| z^i_j \in z_{0...q-1}, z_i \notin z_{0...q-1}, j \in \left[0, q-1\right], i \in \chi\left(j\right) \right\}$ of a reasoning path $z_{0...q-1}$ may derive from multiple \emph{incoming rollbacks}, where  $\chi\left(j\right)$ is the set of rollbacks whose destination is $j$-th thought of this path. As each rollback creates an error experience from one trial of the given question, \emph{incoming rollbacks} lead to a series of experiences 
$A^{\chi\left(\cdot\right)}_{z_{0...q-1}}$. By accumulating an ensemble of trial-and-error reasoning experiences as the in-context learning examples in the prompt, LLM will learn from more experiences to generate the correct next thought $z_{q} \sim f\left(z_{q}|\sI\left(A^{\chi\left(\cdot\right)}_{z_{0...q-1}}, z_{0...q-1}\right)\right)$, as shown in Figure \ref{fig:tr} and two examples of Figure \ref{fig:gsm8k2} and Figure \ref{fig:math2}.

\subsection{Ensembling Solutions}
\label{subsec: ensembling}
TR may create massive final solutions as each adaptive triggered rollback leads to one more new reasoning path toward answering the question. Thus, we directly stop reasoning when there is $K$ number of solutions $\left\{z_{0...T_k}\right\}_{k=1}^K$ obtained. Then, weighted majority voting (W-Voting) will be performed on them for a final solution. Specifically, for the solution $z_{T_k}$, the weight $w_t$ is higher when 1) it has a lower number of \emph{outgoing rollbacks} denoted as $\alpha_{T_k}$, meaning that fewer bad thoughts are identified; and 2) more experiences $\beta_{T_k}=|A^{\chi\left(\cdot\right)}_{z_{0...T_k}}|$ are accumulated along this reasoning path. Eventually, TR outputs the final solution as: $\argmax_{v \in V} \sum_{k=1}^K I\left(v_k=v\right)\left(\beta_{T_k}-\alpha_{T_k}\right)$, where $V$ is the collection of solutions and $v_k$ is the value of $k$-th solution.

\section{Experiments}

\begin{table*}[htbp]
    \footnotesize
    \centering
    \caption{Evaluating the reasoning ability of TR with GPT models under four well-known mathematical problems. TR is specifically evaluated against leading methods such as Faithful-CoT \cite{faithcot-aacl23}, and CSV \cite{csv-iclr24}, each achieving state-of-the-art (SOTA) performance on the \texttt{SVAMP}  and \texttt{MATH}  datasets, respectively. The best results, apart from the SOTA, are in bold. The $5$ shots of CoT examples used by our TR experiments are extracted randomly from the trainset of the same category. }
    \label{table:maincomparision}
    \begin{adjustbox}{max width=\textwidth}
    \begin{tabular} { cccccccccc }
    \toprule

    \multirow{2}{*}{Methods} & \multirow{2}{*}{\makecell{ZeroShot}} & \multicolumn{4}{c}{GPT-4} & \multicolumn{4}{c}{GPT-3.5-turbo}  \\
    \cline{3-10}
        &                               & \texttt{GSM8K} & \texttt{SVAMP} & \texttt{AQuA-RAT} & \texttt{MATH}  
                                        & \texttt{GSM8K} & \texttt{SVAMP} & \texttt{AQuA-RAT} & \texttt{MATH}         \\
    \midrule

    SOTA                & \ding{55}     & 98.7$_{8}$           & 95.4           & 85.04$_{8}$              & 84.32          
                                        & 89.2                 & 84.3           & 60.6$_{8}$              & 40.56\\   
    \midrule
    ZeroShot            & \ding{51}     & 87.1                  & 79.33$^\dag$              & 50.4              & 42.2 
                                        & 76.3                  & 74.8                      & 53.5              & 24.5$^\dag$         \\
    ZeroShot-CoT        & \ding{51}     & 93.1$^\dag$           & 84.67$^\dag$              & 73.2              & 44.7
                                        & 79.6                  & 77.5                      & 53.9              & 30$^\dag$          \\
    CoT             & \ding{55}         & 94.2$_{5}$$^\dag$   & 91.9$_{5}$     & 75.2$_{8}$        & 48.9$_{8}$
                                        & 87.4$_{5, sc_{15}}$   & 83$_5$         & 59.4$_{5}$              & -         \\
    C-CoT     & \ding{55}         & 94.9$_{8}$            & 90.5$_8$       & 77.5$_{8}$        & 50.4$_{8}$
                                        & 82.8$_{8}$            & 81.0$_8$       & 57.4$_{8}$        & 34.1$_{8}$         \\
    PHP+C-CoT & \ding{55}         & 95.5$_{8}$            & 91.9$_8$       & 79.9$_{8}$        & 53.9$_{8}$
                                        & 85.1$_{8}$            & 83.1$_{8}$     & 60.6$_{8}$        & 36.5$_{8}$          \\
    BoT                 & \ding{51}     & 97.1                  & 92.67           & 81.5              & 62.44
                                        & -                     & -              & -                 & \textbf{40.56}    \\
    BoT+CoT       & \ding{55}         & \textbf{98.7$_{8}$}            & \textbf{95$_{8}$}     & 85.04$_{8}$        & 66.33$_{8}$
                                        & -                     & -              & -                 & -     \\
    \midrule
    Chain Reasoning$^\dag$     & \ding{51}     & 89.76           & 80.33           & 74.41              & 45.44
                                               & 76.72           & 71.67            & 47.64              & 26.89         \\
    ToT Reasoning$^\dag$       & \ding{51}     & 90.9            & 84           & 76.38               & 48
                                               & 79.83           & 78.33           & 54.72              & 30.44          \\
    \midrule
    \midrule
    TR$^\dag$                 & \ding{51}     & 94.24           & 89          & 79.92     & 55
                                              & 82.49           & 77.67             &  56.69    &    32.78      \\
    TR + CoT$_5$$^\dag$         & \ding{55}     & 96.06           & 91.33         & 84.25          & 62.56          
                                              & 86.5           & 79.67             & 57.87          & 31.44   \\

    TR + W-Voting$^\dag$                 & \ding{51}     & 96.36           & 93          & \textbf{87.8}     & 71.89
                                        & 85.9   & 82.33          & \textbf{63.39}     & 39.78          \\
    TR + CoT$_5$ + W-Voting$^\dag$         & \ding{55}     & 96.97           & 93.33           & 87.4              & \textbf{72.11}          
                                        & 87.79           & 82.67          & 62.6              & 35.89\\
    \bottomrule
    \end{tabular}
\end{adjustbox}
\vspace{-0.3cm}
\end{table*}

\textbf{Datasets}. We conduct experiments on two streams of tasks. For the mathematical problems, we evaluate the performance of TR on test sets of \texttt{GSM8K}$_{1319}$ \cite{GSM8K-arxiv21}, \texttt{SVAMP}$_{300}$ \cite{SVAMP-arxiv21}, \texttt{AQUA-RAT}$_{254}$ \cite{AQuA-arxiv17}, \texttt{MATH}$_{900}$ \cite{MATH-arxiv21}, \texttt{TheoremQA}$_{400}$ \cite{TheoremQA-emnlp23} datasets, where numerical subscripts indicate sample size. For \texttt{TheoremQA}$_{400}$, we specifically use half of the test set without visual information, leading to $400$ samples. Following ToT \cite{tot-nips23}, we utilize $100$ challenging games of \texttt{\texttt{Game of 24}}. For multi-task reasoning, such as symbolic reasoning, we extract $900$ samples from $56$ categories of \texttt{MMLU} \cite{MMLU-iclr21}, i.e. \texttt{MMLU}$_{900}$. 

\textbf{Large language models}. We utilize GPT-3.5-turbo (gpt-3.5-turbo-16k-0613), GPT-4 (gpt-4-1106-preview) \cite{gpt4-arxiv23} and Llama2 \cite{llama2-arxiv23}, including Llama2-13b (Llama-2-13b-chat-hf) and Llama2-70b (Llama-2-70b-chat-hf) where 1b means one billion parameters. For LLMs with TR, the default settings for temperature and top\_p are $0.7$.

\textbf{Baselines}. Apart from zero-shot prompting \cite{llmszeroshot-nips22}, the comparison approaches include Chain-of-thought (CoT) \cite{cot-nips22}, SC \cite{self-consistency-iclr22} and Complex CoT \cite{complexity-iclr23} (C-CoT), where the subscript $5$ or $8$ indicates the number of shots while the subscript $sc$ denotes the number of sample paths. Also, TR is compared with the related approaches, such as Boosting of thoughts (BoT) \cite{bot-iclr24}, Tree of thoughts (ToT) \cite{tot-nips23}, Cumulative Reasoning (CR) \cite{cr-arxiv23}, and Progressive-Hint Prompting (PHP) \cite{php-arxiv23}. ToT follows the best first search (BFS). The breadth limit of ToT is $6$ while BoT performs $10$ boosting iterations on $15$ binary trees. We also include the state-of-the-art (SOTA) methods, such as CSV \cite{csv-iclr24} that relies on GPT-4 Code Interpreter, on each dataset as an additional comparison. We set $K=8$ for possible early stopping of TR in all experiments.

\textbf{Metrics}. All experiments report the Solve Rate (\%) of the questions. We make LLM explicitly report the solution value after the strings, such as ``The solution is'' and ``The choice is'' in the $z_{0...T}$. Thus, the value is directly extracted and compared with the ground truth. The Interaction Number refers to the frequency at which we must consult the LLM until we receive conclusive responses.

\textbf{Reproducibility}. The results and methods marked with a superscript $^\dag$ are the results we obtained based on the open source platform \textit{llmpebase}. Others without such a tag are collected from existing work, as shown in the Appendix's subsection \ref{subsection: source}.

\subsection{Main Evaluation Results}

\textbf{Adaptive reasoning}. With zero-shot prompting and no pre-defined thought structures, such as chain of Chain Reasoning and the tree of ToT Reasoning, TR allows GPT-3.5-turbo, GPT-4, and Llama2 to self-organize and explore thought structures toward answering the question. Under challenging tasks, LLMs with TR adaptively build complex structures, as shown by examples in the Appendix, by continuously rolling back from thoughts with ``hallucinations''. For simpler tasks, lightweight structures are built by LLMs with TR. As such, with the ability to adjust thoughts and prompt the LLMs with accumulated experience of errors during reasoning, TR achieves a high solving rate and relatively lower resource cost.

\begin{figure*}[h]
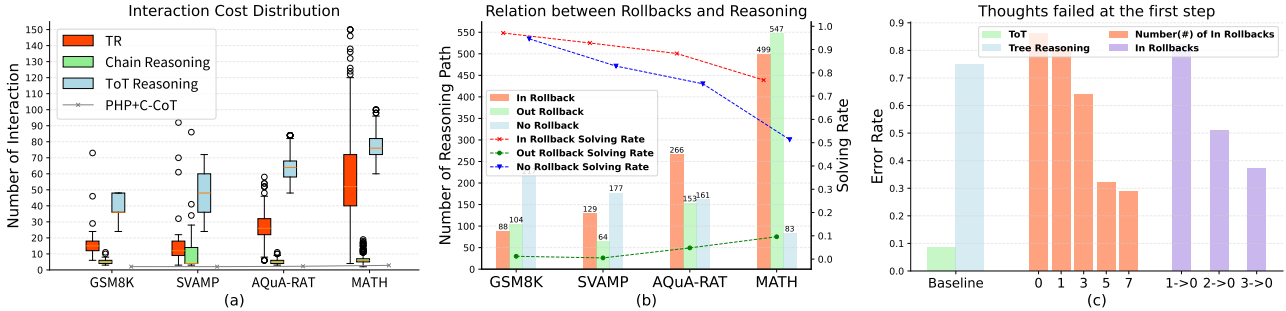

    \centering
    \begin{minipage}{1\textwidth}
    {\includegraphics[width = 0.33\textwidth]{figs/gpt4\_interactions.pdf}}
    {\includegraphics[width = 0.33\textwidth]{figs/gpt4\_rollback.pdf}}
    {\includegraphics[width = 0.33\textwidth]{figs/gpt4\_g24.pdf}}
    \caption{Effectiveness of TR on interaction cost saving and thought revision through the rolling back of thoughts. a). distributions of the interactions required to address problems in four datasets; b). solving rates of three types of reasoning path: ``No Rollback'' --- thoughts receive no rollbacks, ``Out Rollback'' --- rollbacks triggered by mistaken thoughts, and ``In Rollback'' --- thoughts derive from rollbacks; c). the reduction of failure rates due to the rollbacks in the first step of \texttt{\texttt{Game of 24}}, where $\left[0, 1, 3, 5, 7\right]$ denotes the number of rollbacks that cause the first step and $2\rightarrow 0$ means the first step derives from the rollback from the $2$-th step.}
    \label{fig:TRDetails}
    \end{minipage}
    \vspace{-0.3cm}
\end{figure*}

\textbf{Overall comparison}. We show, especially in Table \ref{table:maincomparision}, that compared to existing multi-step reasoning approaches, LLMs with TR achieve the best and the second best solving rate on \texttt{AQuA-RAT} \&\texttt{MATH} and \texttt{GSM8K}  \& \texttt{SVAMP} , respectively. Meanwhile, contrary to the resource-cost SOTA ones, such as BoT, which undertakes reasoning through massive tree thought structures, and CSV, which relies on GPT-4 Code Interpreter, TR yields notable performance by interacting less with relatively simpler LLMs. First, TR surpasses BoT by $6.3$ on \texttt{AQuA-RAT} and $9.45$ on \texttt{MATH} using GPT-4. In particular, TR requires only around $40$ interactions compared to the $500+$ interactions of BoT. Using TR with zero-shot prompting, GPT-4 and GPT-3.5-turbo outperform the ones using few-shot CoT prompting and self-consistency. Under hard math problems, especially MATH, the solving rate of TR is $17.99$ and $3.28$ higher than PHP+C-CoT under GPT-4 and GPT-3.5-turbo, respectively. Second, LLMs with TR adaptively explore thought structures, which is significantly better than pre-defined forward-only Chain Reasoning and ToT Reasoning. After large-scale interactions with LLMs, the performance of the latter two zero-shot prompting methods only approaches the 8 shots Complex CoT. Figure \ref{fig:TRDetails} (a) shows that compared to ToT reasoning, our TR requires one-third or less of the interactions to achieve a new state-of-the-art. Finally, with an average $28$ interactions with GPT-4, TR yields a competitive solving rate $87.56$ on the multi-task dataset MMLU, which contains symbolic reasoning.

We emphasize that TR is more effective in challenging problems, as shown in Table \ref{table:complexreasoning} and Table \ref{table:g24}. In the level 5 difficulty of MATH, GPT-4 with TR is only $2.84$ lower than the CSV that embraces the GPT-4 Code Interpreter as an auxiliary. The solving rate of GPT-4+TR is $4.35$ higher than the current best in \texttt{TheoremQA}. Along with better performance, interaction cost is reduced to an acceptable range by TR. Another observation is that LLMs with TR introduce more interactions in hard problems than the simpler ones. For instance, as shown in Figure \ref{fig:TRDetails} (a) and the first two columns of Table \ref{table:complexreasoning} , the average interaction cost increases to around $60$. Also, in Table \ref{table:g24}, the results of the \texttt{Game of 24} dataset show that GPT-4 with TR requires an average of $32$ interactions to reach a solving rate of $87\%$, which is only $7\%$ lower than the CR$_{2}$ \cite{cr-arxiv23}. This is a remarkable achievement as CR-related approaches rely on human-made demonstrations while TR is zero-shot prompting. Moreover, introducing the CR's demonstrations into the prompt of TR increases the solving rate to $93\%$ while reducing the number of interactions to $24$. In addition, the better performance of TR + CR-Prompt shows that including demonstrations reduces the reliance on majority voting. 

\begin{table}[t]
    \footnotesize
    \centering
    \caption{Evaluating TR with the ZeroShot setting on challenging mathematical problems and multi-task reasoning.  With GPT-4, the existing SOTA zero-shot methods on the level 5 difficulty of the \texttt{MATH}, \texttt{TheoremQA}, and \texttt{MMLU} are from CSV \cite{csv-iclr24}, PoT \cite{pot-tmlr23}, and BoT \cite{bot-iclr24}, respectively. We use $324$ samples out of $1324$ for the \texttt{MATH} dataset. The average interaction number with LLM to solve each problem is reported within $\left(\right)$ behind the number.}
    \label{table:complexreasoning}
    \begin{adjustbox}{max width=\columnwidth}
    \begin{tabular} { cccc }
    \toprule

    Methods                     & \texttt{MATH}-level5$_{324}$  & \texttt{TheoremQA}$_{400}$ & \texttt{MMLU}$_{900}$ \\
    \midrule
    SOTA                        & 55 $\left(3\right)$          & 52.4 $\left(1\right)$          & 93.2 $\left(900+\right)$             \\   
    \midrule
    Llama2-13b$^\dag$           & 2.47 $\left(1\right)$        & 9.3                        & 51.78 $\left(1\right)$       \\
    Llama2-70b$^\dag$           & 8.64 $\left(1\right)$       & 25.5                       & 65.44 $\left(1\right)$       \\
    \midrule
    GPT-4 + ZeroShot-CoT$^\dag$         & 23.46 $\left(1\right)$       & 43.75 $\left(1\right)$      & 82.33 $\left(1\right)$       \\
    \midrule
    GPT-4 + Chain Reasoning$^\dag$      & 22.53  $\left(11\right)$           & 36.5  $\left(9\right)$         & 78.11  $\left(6\right)$            \\
    GPT-4 + ToT Reasoning$^\dag$        & 24.38 $\left(150\right)$           & 38.25  $\left(110\right)$        & 79.22  $\left(70\right)$            \\
    \midrule
    \midrule
    Llama2-70b + TR + W-Voting $^\dag$             & 12.65  $\left(36\right)$     & 29 $\left(34\right)$     & 58.33 $\left(22\right)$              \\
    GPT-3.5-turbo + TR + W-Voting$^\dag$         & 20.06  $\left(38\right)$       & 39.5  $\left(30\right)$    & 70.11  $\left(18\right)$              \\
    GPT-4 + TR$^\dag$                  & 31.48       & 46.25     & 84.67              \\
    GPT-4 + TR + W-Voting$^\dag$                  & 52.16 $\left(62\right)$       & \textbf{56.75} $\left(56\right)$     & 87.56 $\left(28\right)$              \\

    \bottomrule
    \end{tabular}
\end{adjustbox}
\end{table}

\begin{table}[t]
    \footnotesize
    \centering
    \caption{Utilizing TR with GPT-4 achieves competitive performance while maintaining low interaction cost on \texttt{Game of 24} dataset.}
    \label{table:g24}
    \begin{adjustbox}{max width=\columnwidth}
        \begin{tabular}{ccccc}
        \toprule        
        Method                                      &   Solving rate    &   \#Interactions  &   Generate tokens  &   Prompt tokens     \\
        \midrule
        Standard                                    & 7.3               &   1  &   -  &   -        \\
        Standard \scriptsize{(best of 100)}         & 33                &   100 &   1.8k  &   1k            \\
        CoT                                         & 4                 &   1  &   -  &   -        \\
        CoT \scriptsize{(best of 100)}              & 49                &   100  &   6.7k   &   2.2k      \\
        CoT-SC                                      & 9 $_{sc_{100}}$   &   100  &   -  &   -      \\
        ToT\scriptsize{(b = 5)}                     & 74                &   30  &   5.5k  &   1.4k       \\
        CR\scriptsize{(b = 2)}                     & 94                &   27.4  &   -  &   -       \\
        CR\scriptsize{(b = 5)}                     & \textbf{98}                &   29.72  &   -  &   -       \\
        \midrule
        BoT                                         & 83.7              & 724  &   15.8k  &   18.6k        \\
        BoT+CoT$_5$                                 & 84.9              & 543  &   11.2k &    15.5k       \\
        \midrule
        Chain Reasoning$^\dag$                      & 5                 & 3  &   0.14k  &   0.22k          \\
        ToT\scriptsize{(b = 5)} Reasoning$^\dag$    & 25.6              & 36  &   2.82k  &   1.63k         \\
        \midrule
        \midrule
        TR $^\dag$                                  & 70             & \multirow{2}{*}{32}       & \multirow{2}{*}{5.96k}   & \multirow{2}{*}{9.98k}        \\
        TR + W-Voting$^\dag$                        & 87    &       \\
        TR + CR-Prompt $^\dag$                                  & 86             & \multirow{2}{*}{24}       & \multirow{2}{*}{5.03k}   & \multirow{2}{*}{8.1k}        \\

        TR + W-Voting + CR-Prompt$^\dag$                        & 93    &       \\
        \bottomrule
    \end{tabular}
    \end{adjustbox}
\end{table}

\textbf{Effect of the rollback of thoughts}. In Figure \ref{fig:TRDetails} (b) and (c), we specifically present the relation between rollbacks and the solving rate of reasoning paths and the decrease in the failure rates at the first step of the \texttt{Game of 24}. We define a reasoning path $z_{0...T}$ as \textit{In Rollback} if a majority of its thoughts, represented by $\mathbf{Z}^{\chi\left(\cdot\right)}_{z_{0...T}}$ in the subsection \ref{subsec: enhancing}, are derived from \emph{incoming rollbacks}. $z_{0...T}$ is defined as \textit{Out Rollback} if more than two of its thoughts trigger \emph{outgoing rollbacks} and as \textit{No Rollback} if it includes no rollbacks. Figure \ref{fig:TRDetails} (b) presents these three types of reasoning paths TR generates during reasoning in four datasets. As TR allows the error analysis of each rollback to be accumulated in the prompt, as discussed in subsection \ref{subsec: enhancing}, a \textit{In Rollback} path generally benefits from exploiting more experiences during reasoning. Therefore, as Figure \ref{fig:TRDetails} (b) verifies, \textit{In Rollback} paths consistently correspond to a higher solving rate because of prompting LLMs with these trial-and-error experiences. On the contrary, those \textit{Out Rollback} paths have significantly low solving rates because they include more bad thoughts (``hallucinations''), which consequently trigger more rollbacks after being identified by LLMs. Similarly, when the first step of \texttt{Game of 24} derives from the number of $0$ to $5$ \emph{incoming rollbacks}, the failing rate decreases significantly from higher than $0.8$ to lower than $0.3$. The final observation is that longer spans of rolling back are more important for revising thoughts. Figure \ref{fig:TRDetails} (c) shows that the first step caused by a rollback $3\rightarrow 0$ has a higher success rate than the one caused by $2\rightarrow 0$ and $1\rightarrow 0$. Therefore, we argue that \textbf{a rollback, especially generated from more latter reasoning steps (thoughts), contribute more to the thought revision}. This may be because the error analysis brought by rollbacks of later reasoning steps contains more information and is more helpful in improving prompts. 
 
\textbf{Concerns}. First, TR contributes less to the performance enhancement when the LLMs inherently do not have solid ability. Especially in \texttt{AQuA-RAT} and \texttt{MATH} of Table \ref{table:maincomparision}, GPT-3.5-turbo with TR is $2.91$ and $3.25$ higher than PHP+C-CoT, which are significantly lower than these under GPT-4. Likewise, in Table \ref{table:complexreasoning}, the solving rate of Llama2-70b with TR is around 4\% higher than Llama2-70b but costs more interactions. Therefore, the performance of TR depends on LLMs' abilities to perform correct error analysis and understand the experience in the prompt. Second, to address harder problems, LLMs with TR tend to build over-complex thought structures, which generally contain more than $100$ thoughts. The main reason is that a rollback generated by bad thoughts identified by LLMs or a mistaken rollback caused by hallucinations of LLMs leads to one more reasoning path. This appears frequently in challenging tasks; thus, LLMs self-organize a large-scale thought structure toward solutions. We present more visible examples in Figure \ref{fig:introcomplete}, Figure \ref{fig:math1}, and Figure \ref{fig:math2} of the Appendix.

\begin{table*}[ht]
    \footnotesize
    \centering
    \caption{Evaluating the token cost when using GPT-4 with TR to address questions from \texttt{MATH} and \texttt{TheoremQA}. The ratios in the table indicate how many times the token costs of various approaches exceed the token cost of ZeroShot CoT. $\triangle$ represents the increase in the problem-solving rate of various approaches compared to ZeroShot CoT. We also provide the average and standard deviation values, expressed as mean $\pm$ std, for the tokens required to prompt LLMs and those generated by LLMs in addressing a single question from two challenging datasets.}
    \label{table:tokencost}
    \begin{adjustbox}{max width=\textwidth}
    \begin{tabular}{c|ccccc|ccccc}
    \toprule
    \multirow{2}{*}{Method} & \multicolumn{5}{c|}{\texttt{MATH}}     & \multicolumn{5}{c}{\texttt{TheoremQA}}      \\ \cline{2-11} 
                            & \multicolumn{1}{c}{Generate tokens}     & \multicolumn{1}{c|}{Ratios} & \multicolumn{1}{c}{Prompt tokens}         & Ratios  & $\triangle$ & \multicolumn{1}{c}{Generate tokens}    & \multicolumn{1}{c|}{Ratios} & \multicolumn{1}{c}{Prompt tokens}         & Ratios & $\triangle$ \\ \midrule
    ZeroShot CoT            & \multicolumn{1}{c}{212.3 $\pm$ 152.6}   & \multicolumn{1}{c|}{1}      & \multicolumn{1}{c}{110.3 $\pm$ 25.5}      & 1      & 0 & \multicolumn{1}{c|}{240.8 $\pm$ 100.5}  & \multicolumn{1}{c|}{1}      & \multicolumn{1}{c}{121.2 $\pm$ 43.1}      & 1       & 0 \\ 
    CoT$_{8}$               & \multicolumn{1}{c}{441.3 $\pm$ 310.7}   & \multicolumn{1}{c|}{2.08}   & \multicolumn{1}{c}{4611.4 $\pm$ 1952.2}   & 41.8   & 4.2 & \multicolumn{1}{c|}{-}                  & \multicolumn{1}{c|}{-}      & \multicolumn{1}{c}{-}                     & -       & -\\ 
    Chain Reasoning         & \multicolumn{1}{c}{603.2 $\pm$ 244.7}   & \multicolumn{1}{c|}{2.84}   & \multicolumn{1}{c}{2657.5 $\pm$ 1187}     & 24.1   & 0.74 & \multicolumn{1}{c|}{976.3 $\pm$ 1056.4} & \multicolumn{1}{c|}{4.05}   & \multicolumn{1}{c}{4614.6 $\pm$ 7632.1}   & 38.07   & -7.25\\ 
    ToT Reasoning           & \multicolumn{1}{c}{1569.8 $\pm$ 474.7}  & \multicolumn{1}{c|}{7.39}   & \multicolumn{1}{c}{9963.7 $\pm$ 4487.5}   & 90.33  & 3.3 & \multicolumn{1}{c|}{1886 $\pm$ 982.5}   & \multicolumn{1}{c|}{7.83}   & \multicolumn{1}{c}{11982 $\pm$ 8872.2}    & 98.86   & -5.5\\ \hline\hline
    TR + W-Voting            & \multicolumn{1}{c}{7484.9 $\pm$ 5873.9} & \multicolumn{1}{c|}{\textbf{35.26}}  & \multicolumn{1}{c}{46904.9 $\pm$ 37980.8} & 425.25 & 27.19 & \multicolumn{1}{c|}{6762.9 $\pm$ 6513}  & \multicolumn{1}{c|}{28.09}  & \multicolumn{1}{c}{43444.8 $\pm$ 49391.2} & \textbf{358.46} & 13 \\ 
    \bottomrule
    \end{tabular}
\end{adjustbox}
\end{table*}

\subsection{Main Insights}

The notable performance enhancement of TR in terms of both solving rate and interaction cost shows the insight that adaptive adjusting thoughts supported by the rollback of thoughts during reasoning is core to the success of LLMs in complex mathematical reasoning. In addition, we can gain three more insights. 

\textbf{Experience accumulation of error analysis from intermediate thoughts is better than that obtained by analyzing the whole reasoning path}. Existing work \cite{selfcorrection-iclr24} pointed out that LLMs are unable to revise reasoning based on the outcome analysis, which gives feedback on the final reasoning. Thus, BoT \cite{bot-iclr24}, which relies on outcome analysis, had to embrace more careful-selected outcome analysis to prompt LLMs. Our TR opens a new direction of relying on process analysis, which provides error analysis for each intermediate reasoning step (rollback-by-rollback verification), to revise thought adaptively during reasoning. Besides, with the rollback of thought, outcome analysis is a special case of process analysis when analysis is used not to re-do reasoning but to adjust previous thought to create a correct reasoning path. 

\textbf{Experience-guided Solution Ensemble is critical to the effectiveness of trial-and-error reasoning}. After stopping reasoning, LLMs with TR yield $K$ reasoning paths due to the adaptive exploration. Each reasoning path caused by one rollback of TR can be regarded as a trial for addressing the problem. When LLMs frequently make mistakes and occur ``hallucinations'', the solution obtained in any trial may not be correct. Since TR exploits the error analysis of each \emph{incoming rollback} as experience to prompt LLMs, a solution from the reasoning path with more \emph{incoming rollback} is more acceptable. Therefore, we should ensemble these solutions by filtering out ones with limited experiences or many bad thoughts. As shown by comparisons between TR and TR + W-Voting in Table \ref{table:maincomparision}, Table \ref{table:complexreasoning}, Table \ref{table:g24} and Figure \ref{fig:TRDetails} (b), such an experience-guided solution ensemble is critical. 

\textbf{Weak LLMs may not identify multiple targets mentioned in the prompt}. Including CoT examples in the prompt improves the solving rate of LLMs with TR, as shown in the GPT-4 column of Table \ref{table:maincomparision}. However, for the more challenging \texttt{AQUA-RAT} and \texttt{MATH} datasets in the GPT-3.5-turbo column, adding CoT causes a significant performance decrease. We argue that it may be hard for the weak LLMs to understand and distinguish instructions with different targets in the prompt. For example, the instruction of CoT examples emphasizes how to follow demonstrations, while the prompt with experiences in TR focuses on how to avoid given errors. Weak LLMs may not benefit from the enhanced prompt containing these two different guidances, especially when the reasoning is complex. 

\subsection{Token Cost Analysis}

As shown in Table~\ref{table:tokencost}, when addressing questions from challenging datasets, the token cost of GPT-4 with TR is significantly higher than baseline approaches. Specifically, on the \texttt{MATH} dataset, the TR approach, on average, generates 35.26 times more tokens and requires $425.25$ times more prompt tokens than the ZeroShot CoT. The corresponding ratios in the \texttt{TheoremQA} dataset are $28.09$ and $358.46$. This resource-intensive nature of our proposed TR derives from continuously identifying the errors and appending the prompt with error analysis during reasoning. 

However, these additional operations and the high token cost are necessary as hallucinations of LLMs frequently appear. First, since numerous erroneous thoughts are generated during reasoning, consistently identifying and revising them is crucial to ensuring the correctness of the answer. Second, in many cases, the error analysis of LLMs is invalid and even incorrect due to the inevitable hallucinations. Thus, accumulating error analysis derived from different reasoning paths decreases the negative impact of the flawed analysis on thought revisions. Third, many reasoning paths of the thought structure derive from rollbacks triggered by erroneous feedback from LLMs. Since mistaken rollbacks cannot be identified, the TR approach retains all generated paths and ultimately employs majority voting to enhance reliability.

Therefore, we conclude that there is a trade-off between the high token cost and the problem-solving rate. On the one hand, since the TR approach requires many tokens to address a single question, its application may be limited for users with insufficient resources. On the other hand, compared to zero-shot GPT-4, GPT-4 with TR gains 27.19\% and 13\% improvements in solving rates on \texttt{MATH} and \texttt{TheoremQA} datasets. When users prioritize problem-solving rates, integrating GPT-4 with the TR approach ensures its applicability in many challenging scenarios.

\section{Concluding Remarks}
In this paper, we proposed Thought Rollback (TR), an effective reasoning framework supported by the rollback of thoughts that allows LLMs to perform adaptive reasoning to solve challenging problems. Without relying on human annotations and specific thought structure designs for reasoning, LLMs with TR can progressively self-organize and revise thoughts based on trial-and-error experiences until reaching a correct solution for various tasks. The \textit{rollback controller} and \textit{prompt enhancer}, together with the experience-guided weights majority voting, make TR achieve the state-of-the-art solving rate in many mathematical and multi-task reasoning datasets while maintaining a lower cost than the alternative leading approaches. We hope this work could shed light on the adaptive reasoning in LLMs toward addressing challenging tasks, especially when mathematical problems are involved. 

\section*{Impact Statement}
Large language models (LLMs) can break a complex task into manageable subproblems and solve them through step-by-step reasoning. TR, a lightweight framework, guarantees the reliability of LLMs' multi-step reasoning under hallucinations, thus extending their applications to a wider range of tasks. Furthermore, compared to the BOT that performs outcome analysis, TR built upon prompting the LLMs with feedback from process analysis is more effective and significantly reduces the interaction cost. This may open a research direction emphasizing the importance of exploiting feedback during the step-by-step reasoning of LLMs. In addition, the plug-and-play nature of TR allows other approaches, such as CR, to involve the thought rollback mechanism to improve performance further. Ultimately, thanks to the expansion in the context window of LLMs and the decreasing token prices, the token cost of TR may not be a major concern.

\nocite{langley00}

\bibliography{main}
\bibliographystyle{main}

\newpage

\appendix

\onecolumn

\section{Supplement to Figures of the Main Paper}

\begin{figure*}[ht]
    \centering
    \includegraphics[width=\textwidth]{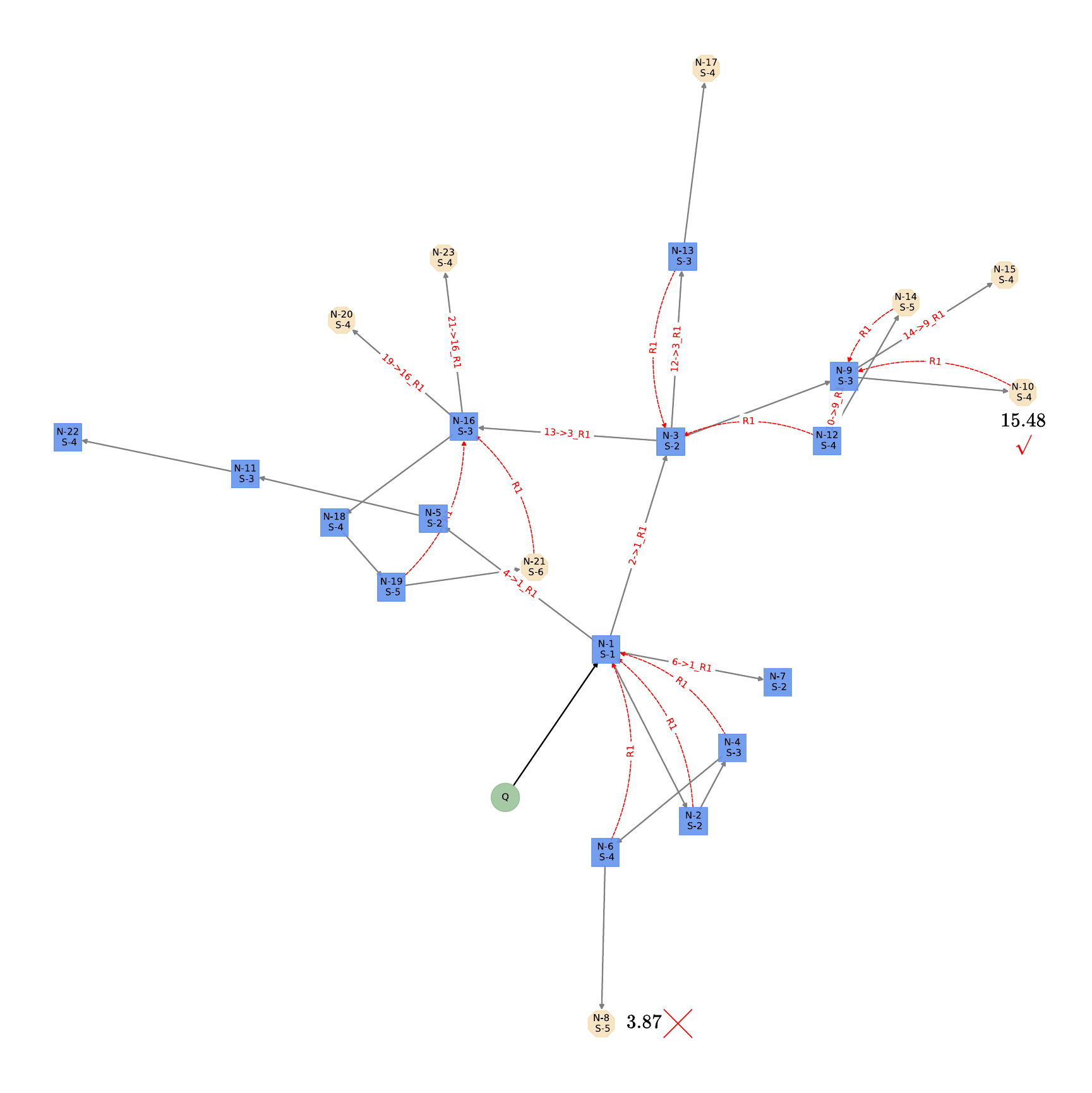}
    \caption{Complete thought structure of Figure \ref{fig:intro} (c) built by GPT-4 with TR for the question from \texttt{MATH} dataset \cite{MATH-arxiv21}. This structure contains $23$ nodes, i.e. $23$ thoughts and leads to $K=8$ reasoning paths towards solutions. It is plotted by based on the NetworkX lib \cite{netx-08} under the ``fdp'' layout.}
    \label{fig:introcomplete}
\end{figure*}

Figure \ref{fig:introcomplete} is the full version of the partial one shown in Figure \ref{fig:intro}. This shows a complex thought structure built by GPT-4 with TR to address a challenging question from \texttt{MATH} dataset. It shows that GPT-4 with TR tends to build a large-scale thought structure by iteratively rolling back thoughts during reasoning. Figure \ref{fig:introcomplete} demonstrates the details about how TR exploits \textit{rollback controller} and \textit{prompt enhancer} to create a new and correct reasoning path after analyzing the bad thoughts. 

\begin{figure*}[ht]
    \centering
    \includegraphics[width=\textwidth]{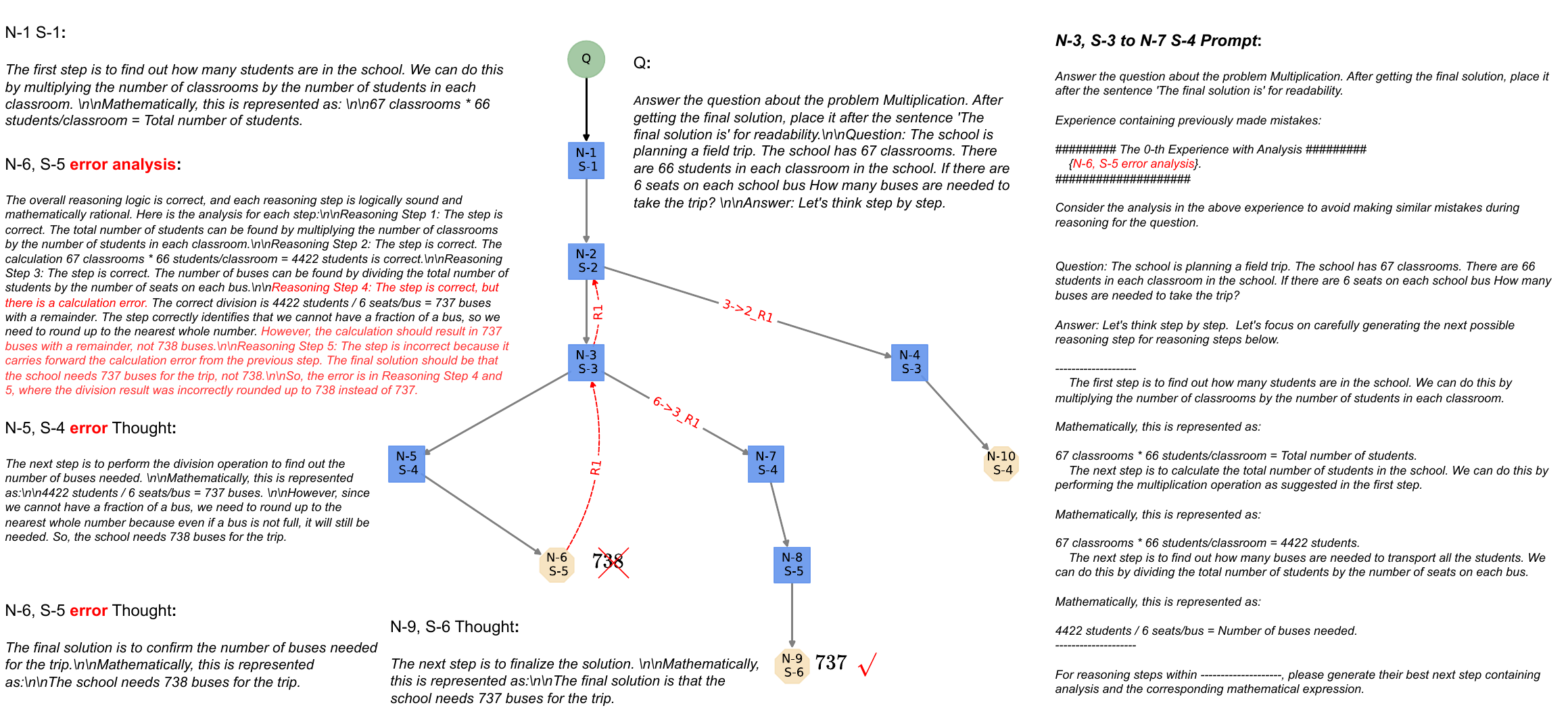}
    \caption{Complete thought structure of Figure \ref{fig:tr} built by GPT-4 with TR for the question from \texttt{SVAMP} dataset \cite{SVAMP-arxiv21}. This structure contains $10$ nodes, i.e. $10$ thoughts and leads to $K=3$ reasoning paths towards solutions. It is plotted by based on the NetworkX lib \cite{netx-08} under the ``dot'' layout.}
    \label{fig:fig2example}
\end{figure*}

Specifically, from the details presented in Figure \ref{fig:fig2example}, we can observe what TR shown in Figure \ref{fig:tr} does after reaching the thought. GPT-4 with TR starts from generating a first thought $N-1 S-1$ based on a simple and zero-shot prompt $Q$. Then, \textbf{Rollback of thoughts} follow the process in Figure \ref{fig:tr}. First, \textit{rollback controller} analyzes the reasoning path $N-1 \rightarrow N-2 \rightarrow N-3 \rightarrow N-5 \rightarrow N-6$ to output the error analysis presented in ``N-6, S-5 \textcolor{red}{error analysis}:''. This analysis shows that reasoning steps $N-5$ $S-4$ and $N-6$ $S-5$ are bad thoughts. According to our discussion in subsection \ref{subsec: rollback}, \textit{rollback controller} allows the LLM to roll back to the thought $N-3 S-3$, which is one step before the first bad thought $N-5$ $S-4$. Then, \textit{prompt enhancer} accumulates the error analysis as experience in the prompt, as shown in ``\textbf{N-3, S-3 to N-7 S-4 Prompt}'' which includes the ``\#\#\#\# The 0-th Experience with Analysis \#\#\#\#''. As a result, by avoiding making similar mistakes mentioned by experience, LLM is able to generate a new thought $N-7$ $S-4$ from the chosen thought $N-3$ $S-3$. Therefore, ``hallucinations'' that occur in thought or analysis of LLM may not influence reasoning due to the continuous thought revision guaranteed by the iterative rollbacks during reasoning. As can be seen from $N-9$ $S-6$, the final solution is revised to be a correct answer $737$. Additionally, we can also observe that each rollback will lead to a new reasoning path with the experience from the corresponding error analysis. Thus, two rollbacks $N-6 \rightarrow N-3$ and $N-3 \rightarrow N-2$ of GPT-4 with TR adaptively create two new reasoning paths.

\section{Reproducibility of Thought Rollback Framework}

\subsection{Source Code}
One can also access the source code under the \textit{examples/ThoughtRollback} folder in the \textit{code}. The implementation is based on the \textit{llmpebase} library. The code is written in Python and imports the datasets from Hugging Face to build PyTorch's data loader.

Also, the source code for Chain Of Thought, Chain Reasoning, ToT reasoning, and GoT reasoning that we mention in experiments are available in the \textit{examples/ChainOfThought}, \textit{examples/ChainReasoning}, \textit{examples/TreeReasoning}, and \textit{examples/GraphReasoning} folders, respectively.

All configuration files used to conduct the experiments are provided in the \textit{configs/} folder in the \textit{code}. About how to run, please read the README.md under the \textit{examples/ThoughtRollback} folder.

\subsection{Locations of Generated Thoughts and Reasoning Details}
\begin{figure}[ht]
    \centering
    \includegraphics[width=\textwidth, height=0.5\textheight, keepaspectratio]{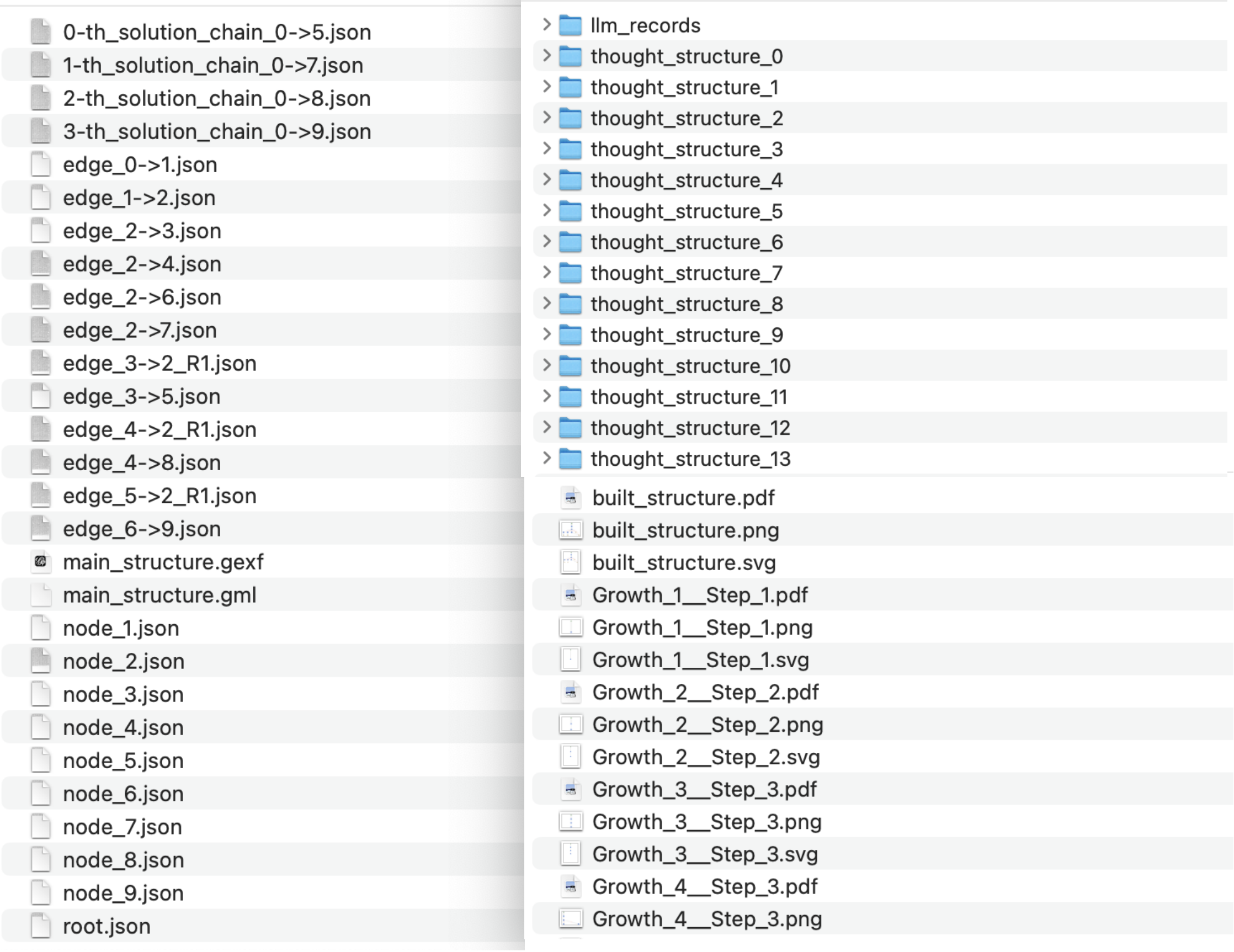}
    \caption{Illustrating of all results generated by LLMs with TR. The left sub-figure presents the details of the generated thought structure. The upper right sub-figure shows the files of obtained results, while the lower right sub-figure presents the visualizations.}
    \label{fig:locations}
\end{figure}

Our code will automatically save the generated thoughts and reasoning details under one folder of \textit{LLMPE} in the root directory. The direct results are placed under the \textit{LLMPE/results} while the corresponding visible thought structures are stored in\textit{LLMPE/visualizations}. Then, their sub-folder name will represent the setting of the configuration, such as ``TRReasoning\_\_gpt-4\_\_zeroshot\_cot\_\_MATH'', where ``TRReasoning'' is the name of our Thought Rollback. Eventually, as shown in Figure \ref{fig:locations}, you can access the sample by index in \textit{thought\_structure\_*} while reading the results in \textit{llm\_records}. The thought structure for reasoning will be saved in the \textit{.json} format, and the visualizations will be in the \textit{.pdf} format.

\subsection{Prompts}
\label{sec: basePrompt}

This subsection presents the basic prompts used in our implementation of Thought Rollback framework. 

\textbf{System prompt for thought generation}: \textit{You possess expertise in solving mathematical problems through a systematic, step-by-step reasoning process during which you are dedicated to preventing repeating any errors analyzed in experiences. Your objective is to address the question using a series of reasoning steps delivered in multiple responses, with each response containing one reasoning step. It is crucial to avoid repeating errors mentioned in the given experiences. Begin by reading the provided reasoning steps and then proceed to generate the most appropriate next step in the response, ensuring that the logical progression steadily leads towards a solution.}

\textbf{System prompt for reasoning analysis}: \textit{You are a mathematician specializing in checking and analyzing the reasoning process containing multiple intermediate reasoning steps proposed to address a math question. Please check the correctness of the overall reasoning logic and each reasoning step regarding mathematical logic and rationality.}

\textbf{Prompt $\sI$ for the next thought generation}: 

\textit{Answer the question about the problem \{Problem Name\}. After getting the final solution, place it after the sentence 'The final solution is' for readability.$\backslash n$$\backslash n$Experience containing previously made mistakes:$\backslash n$$\backslash n$\#\#\#\#\#\#\#\#\#\{Experiences\}\#\#\#\#\#\#\#\#\#$\backslash n$$\backslash n$Consider the analysis in the above experience to avoid making similar mistakes during reasoning for the question.$\backslash n$$\backslash n$$\backslash n$Question: \{Question\} $\backslash n$$\backslash n$Answer: Let's think step by step.  Let's focus on carefully generating the next possible reasoning step for reasoning steps below.$\backslash n$$\backslash n$\-\-\-\-\-\-\-\-\-\-\-\-\-\-\-\-$\backslash n$\{Existing Reasoning Steps\}$\backslash n$\-\-\-\-\-\-\-\-\-\-\-\-\-\-\-\-$\backslash n$$\backslash n$For reasoning steps within \-\-\-\-\-\-\-\-\-\-\-\-\-\-\-\-, please generate their best next step containing analysis and the corresponding mathematical expression.}

where the \textit{\{Problem Name\}} contains what question to solve, such as ``Multiplication'', \textit{\{Experiences\}} present the accumulated experiences $A^{\chi\left(\cdot\right)}_{z_{0...q-1}}$ discussed in subsection \ref{subsec: enhancing}, \textit{\{Question\}} is the given question description and finally \textit{\{Experiences\}} is a placeholder to be replaced by the preceding chain of thoughts $z_{0..,n-1}$.

\textbf{Prompt $\sI_R$ for the error analysis}: 

\textit{Analyze the reasoning steps proposed for the question about the problem \{Problem Name\}. $\backslash n$Question: \{Question\} $\backslash n$ Toward addressing the given question, below is a reasoning process containing \{Number of Steps\} steps: $\backslash n$$\backslash n$\-\-\-\-\-\-\-\-\-\-\-\-\-\-\-\-$\backslash n$ \{Existing Reasoning Steps\} $\backslash n$\-\-\-\-\-\-\-\-\-\-\-\-\-\-\-\-$\backslash n$$\backslash n$Double-check the reasoning process within \-\-\-\-\-\-\-\-\-\-\-\-\-\-\-\-, please analyze its overall and each step's correctness by checking whether they are mathematical logic and rationality. Please report an error when any step does not contain a clear mathematical expression. Output empty string when no steps are given.$\backslash n$}

where the \textit{\{Number of Steps\}} presents the number of current reasoning steps, i.e., $n$ for a reasoning path $z_{0..,n}$.

\subsection{Basic Engineering Settings for TR}

During implementation, we set the upper bound $U$ of the \emph{incoming rollback} for one thought to be $3$, making no more than $3$ rollbacks use this thought as the destination. Besides, we utilize the depth-first search algorithm to find the current growing thought. As mentioned in the subsection \ref{subsec: ensembling}, the reasoning path with more \emph{incoming rollback} is more important. Therefore, for this depth-first search algorithm, we assign higher priority to the reasoning path with a larger number of \emph{incoming rollbacks}. As LLMs with TR first generate subsequent thoughts for these reasoning paths, the corresponding solution will be explored in advance. Such a mechanism increases the possibility of getting better answers. 

Apart from these basic settings, our implementation also includes some engineering tricks. First, during reasoning, once a reasoning path causes \emph{outgoing rollbacks} more than $5$ times, it will be ignored in the subsequent reasoning. Second, we do not allow one thought to cause more than $3$ \emph{outgoing rollbacks} to avoid the case that LLMs repeatedly identify a bad thought and trigger rollbacks for it. Third, to increase the speed of reasoning, after noticing that different reasoning paths are independent of each other, we achieve parallel running, which makes LLM generate thoughts for all reasoning paths simultaneously. Therefore, once a new reasoning path is created, one process will be created to make LLM work on the path without blocking others. This can increase the reasoning speed from the scale of minutes to seconds.

\section{Discussion: Insights Gained from the TR Framework}

We argue that the outstanding performance of TR is attributed to three insights. 

First, as mentioned by the work \cite{letverify-iclr24}, compared to outcome supervision, which provides feedback for a final result, process supervision, which provides feedback for each intermediate reasoning step, is more important to guarantee the reliable reasoning of LLMs. In summary, process supervision significantly outperforms outcome supervision. Actually, TR roughly belongs to process supervision because it performs rollback-by-rollback error analysis during reasoning and continues to revise thoughts based on the accumulation of experience. Thus, the reasoning of LLMs can be adjusted adaptively during reasoning rather than after the results are obtained. However, BoT \cite{bot-iclr24} belongs to outcome supervision as it only collects error analysis after reasoning.

Second, the work \cite{selfcorrection-iclr24} figured out that the outcome analysis is sometimes invalid as LLMs cannot use feedback to revise thoughts to improve the solving rate. As discussed in subsection 3.2 of our paper, this may be because when there are many errors in the intermediate reasoning steps, capturing the source mistake and performing useful analysis is challenging. Thus, BoT \cite{bot-iclr24} has to do many tree ensembles and design a complex boosting mechanism to mine effective error analysis to drive LLM's reasoning revision. On the contrary, TR performs thought rollback triggered by continuous error analysis during reasoning. Thus, once the error is identified, LLMs can directly revise or adjust the reasoning based on timely accumulated experience in the prompt.

Third, prompting LLMs with a long input context may cause the degradation of reasoning performance. As pointed out by the work \cite{lostinlong-acl24}, LLMs do not fairly utilize all the information in the prompt but focus more on the content at the beginning or end of the input context. For BoT, the prompt tends to become extremely long due to the continuous collection of generated reasoning steps and their error analysis over iterations, particularly since the error analysis focuses on the whole reasoning chain. For instance, in the $10$-th iteration, the prompt contains ten long contents, each containing all reasoning steps and step-wise analysis. Some of them may not even be correct. However, our TR accumulates experience during reasoning; thus, each experience only contains analysis on very few intermediate steps, generally leading to a quick revision after the rollback. For instance, LLMs with TR can easily generate a correct second step based on the experience ``Reasoning step 1: ... correct. Reasoning step 2: ... wrong because....''. However, the BoT' experience ``Reasoning step1: ....correct. Reasoning step 2: ... wrong because.... . Reasoning step 3: ... wrong because..... Reasoning step 4: ... wrong because.... . Reasoning step 5: ... wrong because....'' may just mislead the LLM.

Finally, regarding the time complexity of the TR framework, LLMs with TR incrementally construct the thought structure from \#node $0$ to \#node $n$. By focusing on the leading term and disregarding constants and lower order terms, the worst-case time complexity of TR is determined to be $\mathcal{O}(n^2)$.

\subsection{Source of Experimental Results}
\label{subsection: source}

In Table \ref{table:gsm8kresult} and Table \ref{table:svampresult}, we collect experimental results of GPT-4 and GPT-3.5-turbo on various settings. We especially show the corresponding work that reports the results. 

\begin{table*}[h]
    \footnotesize
    \centering
    \caption{Source of experiment results of \texttt{GSM8K}. Methods are CSV \cite{csv-iclr24}, PHP \cite{php-arxiv23} with GPT-3.5-turbo-, Model Selection (MS) \cite{automs-emnlp23}, PAL \cite{pal-icml23}, PLAY \cite{roleplay-arxiv23}, Faithful \cite{faithcot-aacl23}, Exps \cite{gpt4exps-arxiv23}, PoT \cite{pot-tmlr23}, IG \cite{isgeneral-emnlp23}, BoT \cite{bot-iclr24}.}
    \label{table:gsm8kresult}
    \begin{adjustbox}{max width=\textwidth}
    \begin{tabular} { ccccccccccccc }
    \toprule

    \multirow{2}{*}{Source} & \multirow{2}{*}{} & \multicolumn{5}{c}{GPT-4} & \multicolumn{5}{c}{GPT-3.5-turbo}  
    \\
    \cline{3-12}
                                            &  & ZeroShot  & FewShot & ZeroShot-CoT & CoT  & Complex-CoT 
                                               & ZeroShot  & FewShot & ZeroShot-CoT & CoT  & Complex-CoT    \\
    \midrule

    \multirow{3}{*}{CSV}  &  & -         & -              & -     & 92.0$_{5}$   & -               
                             & -         & 57.1$_{5}$     & -     & -            & -     \\

                      & Code & 92.9, 94.9$_{sc_5}$      & -       & -           & -   & -               
                             & -                        & -       & -           & -   & -      \\
                  & Code+CSV & 94.5, 97$_{sc_5}$        & -       & -           & -   & -               
                             & -                        & -       & -           & -   & -    \\
    \cline{2-12}
    \multirow{2}{*}{PHP} &   & -         & -       & -            & -    & 94.9$_{8}$               
                             & -         & -       & -            & -    & 82.8$_{8}$     \\
                   & +PHP    & -         & -       & -            & -    & 95.5$_{8}$               
                             & -         & -       & -            & -    & 85.1$_{8}$     \\
    \cline{2-12}                                          
    \multirow{3}{*}{MS} &  & -      & -        & -    & 94.6$_{5}$, 95.6$_{5, sc_5}$, 95.6$_{5, sc_{15}}$    & -               
                           & -      & -        & -    & 80.8$_{5}$, 85.4$_{5, sc_5}$, 87.4$_{5, sc_{15}}$    & -     \\
                   & PAL   & -      & -        & -    & 94.0$_{5}$, 94.7$_{5, sc_5}$, 95.5$_{5, sc_{15}}$    & -               
                           & -      & -        & -    & 79.2$_{5}$, 80.9$_{5, sc_5}$, 82.4$_{5, sc_{15}}$    & -     \\   
                   & Ours  & -      & -        & -    & 95.6$_{5}$, 96.5$_{5, sc_5}$, 96.8$_{5, sc_{15}}$    & -               
                           & -      & -        & -    & 82.6$_{5}$, 88.2$_{5, sc_5}$, 89.2$_{5, sc_{15}}$    & -     \\  
    \cline{2-12}                         
    \multirow{2}{*}{PLAY} &  & -         & -        & -    & -       & -               
                             & 76.0      & -        & 79.6 & 76.9    & -     \\  
                     & Role  & -         & -        & -    & -       & -               
                             & 78.2      & -        & -    & -    & -     \\  
    \cline{2-12}                        
    \multirow{2}{*}{Faithful} &  & 46.9  & -        & -    & 64.9$_{8}$       & -               
                                 & -     & -        & -    & -       & -     \\  
                     & Faithful  & -     & -        & -    & 95$_{8}$       & -               
                                 & -     & -        & -    & -    & -     \\
    \cline{2-12}
    Exps                      &   & 87.1  & -        & -    & -       & -               
                                  & -     & -        & -    & -       & -     \\   
    \cline{2-12}
    PoT*                       &   & -     & -        & -    & -       & -               
                                  & 76.3  & -        & -    & -       & -     \\   
    \cline{2-12}
    IG                        &   & -     & -        & -    & -       & -               
                                  & 23.8  & -        & 78.9    & -       & -     \\   
    \cline{2-12}
    \multirow{2}{*}{BoT}      &  & 87.1  & -        & 89.6     & 92$_{8}$       & 94.9               
                                 & -     & -        & -        & -              & -     \\  
                         & BoT   & -     & -        & 97.1     & 98.7$_{8}$       & -               
                                 & -     & -        & -        & -     & -     \\  
    \bottomrule
    \end{tabular}
\end{adjustbox}
\end{table*}

\begin{table*}[h]
    \footnotesize
    \centering
    \caption{Source of experiment results of \texttt{SVAMP}, \texttt{AQuA-RAT}, \texttt{MATH}, \texttt{TheoremQA} and \texttt{MMLU}. They are CSV \cite{csv-iclr24}, PHP \cite{php-arxiv23}, Model Selection (MS) \cite{automs-emnlp23}, PAL \cite{pal-icml23}, PLAY \cite{roleplay-arxiv23}, Faithful \cite{faithcot-aacl23}, Exps \cite{gpt4exps-arxiv23}, PoT \cite{pot-tmlr23}, IG \cite{isgeneral-emnlp23}, BoT \cite{bot-iclr24}, CR \cite{cr-arxiv23}, TheoremQA \cite{TheoremQA-emnlp23} and GPT4-report \cite{gpt4-arxiv23}.}
    \label{table:svampresult}
    \begin{adjustbox}{max width=\textwidth}
    \begin{tabular} { ccccccccccccc }
    \toprule
    \multicolumn{12}{c}{\texttt{SVAMP}}  \\
    \multirow{2}{*}{Source} & \multirow{2}{*}{} & \multicolumn{5}{c}{GPT-4} & \multicolumn{5}{c}{GPT-3.5-turbo}  
    \\
    \cline{3-12}
                                            &  & ZeroShot  & FewShot & ZeroShot-CoT & CoT  & Complex-CoT 
                                               & ZeroShot  & FewShot & ZeroShot-CoT & CoT  & Complex-CoT    \\
    \midrule

    \multirow{2}{*}{PHP} &   & -         & -       & -            & -    & 90.5$_{8}$               
                             & -         & -       & -            & -    & 81.0$_{8}$     \\
                   & +PHP    & -         & -       & -            & -    & 91.9$_{8}$               
                             & -         & -       & -            & -    & 83.1$_{8}$     \\
    \cline{2-12}                                          
    \multirow{3}{*}{MS} &  & -      & -        & -    & 91.9$_{5}$    & -               
                           & -      & -        & -    & 83$_{5}$    & -     \\
                   & PAL   & -      & -        & -    & 92.2$_{5}$    & -               
                           & -      & -        & -    & 80.3$_{5}$    & -     \\   
                   & Ours  & -      & -        & -    & 93.7$_{5}$    & -               
                           & -      & -        & -    & 84.3$_{5}$    & -     \\  
    \cline{2-12}                         
    \multirow{2}{*}{PLAY} &  & -         & -        & -    & -       & -               
                             & 75.3      & -        & 76.3 & 82.2    & -     \\  
                     & Role  & -         & -        & -    & -       & -               
                             & 83.8      & -        & -    & -    & -     \\  
    \cline{2-12}                        
    \multirow{2}{*}{Faithful} &  & -  & 88.4        & -    & 80$_{8}$       & -               
                                 & -     & -        & -    & -       & -     \\  
                     & Faithful  & -     & -        & -    & 95.4$_{8}$       & -               
                                 & -     & -        & -    & -    & -     \\
    \cline{2-12}
    PoT*                       &   & -  & -        & -    & -       & -               
                                  & 88.2     & -        & -    & -       & -     \\   
    \cline{2-12}
    IG                        &   & -     & -        & -        & -       & -               
                                  & 74.8  & -        & 77.5     & -       & -     \\   
    \cline{2-12}
    \multirow{2}{*}{BoT}      &  & 68.7  & -        & 74.3     & 77.6$_{8}$       & 90.5$_{8}$               
                                 & -     & -        & -        & -              & -     \\  
                         & BoT   & -     & -        & 92.7       & 94.9$_{8}$       & -               
                                 & -     & -        & -        & -     & -     \\  
    \bottomrule
    \bottomrule
    \multicolumn{12}{c}{\texttt{AQuA-RAT}}  \\
    \multirow{2}{*}{Source} & \multirow{2}{*}{} & \multicolumn{5}{c}{GPT-4} & \multicolumn{5}{c}{GPT-3.5-turbo}  
    \\
    \cline{3-12}
                                            &  & ZeroShot  & FewShot & ZeroShot-CoT & CoT  & Complex-CoT 
                                               & ZeroShot  & FewShot & ZeroShot-CoT & CoT  & Complex-CoT    \\
    \midrule

    \multirow{2}{*}{PHP} &   & -         & -       & -            & -    & 77.5$_{8}$               
                             & -         & -       & -            & -    & 57.4$_{8}$     \\
                   & +PHP    & -         & -       & -            & -    & 79.9$_{8}$               
                             & -         & -       & -            & -    & 60.6$_{8}$     \\
    \cline{2-12}                                                                 
    \multirow{2}{*}{PLAY} &  & -         & -        & -    & -       & -               
                             & 53.5      & -        & 53.9 & 59.4    & -     \\  
                     & Role  & -         & -        & -    & -       & -               
                             & 63.8      & -        & -    & -    & -     \\  
    \cline{2-12}                        
    \multirow{2}{*}{Faithful} &  & 50.4  & -        & -    & 75.2$_{8}$       & -               
                                 & -     & -        & -    & -       & -     \\  
                     & Faithful  & -     & -        & -    & 73.6$_{8}$       & -               
                                 & -     & -        & -    & -    & -     \\
    \cline{2-12}
    PoT*                       &   & -     & -        & -    & 72.4       & -               
                                  & -  & -        & -    & -       & -     \\   
    \cline{2-12}
    IG                        &   & -     & -        & -    & -       & -               
                                  & 28.0  & -        & 53.5 & -       & -     \\   
    \cline{2-12}
    \multirow{2}{*}{BoT}      &  & 40.6  & -        & 73.2     & 74$_{8}$       & 77.5               
                                 & -     & -        & -        & -              & -     \\  
                         & BoT   & -     & -        & 81.4     & 84.9$_{8}$       & -               
                                 & -     & -        & -        & -     & -     \\  
        \bottomrule
        \bottomrule     
    \multicolumn{12}{c}{\texttt{MATH}}  \\
    \multirow{2}{*}{Source} & \multirow{2}{*}{} & \multicolumn{5}{c}{GPT-4} & \multicolumn{5}{c}{GPT-3.5-turbo}  
    \\
    \cline{3-12}
                                            &  & ZeroShot  & FewShot & ZeroShot-CoT & CoT  & Complex-CoT 
                                               & ZeroShot  & FewShot & ZeroShot-CoT & CoT  & Complex-CoT    \\
    \midrule

    \multirow{3}{*}{CSV}  &  & 42.2         & -              & -     & -            & 50.36$_{8}$               
                             & -            & -              & -     & -  & 34.12$_{8}$     \\

                      & Code & 69.69, 79.88$_{sc_{16}}$      & -       & -           & -   & -               
                             & -                        & -       & -           & -   & -      \\
                  & Code+CSV & 73.54, 83.54$_{sc_{16}}$, 84.32$_{sc^{vw}_{16}}$        & -       & -           & -   & -               
                             & -                        & -       & -           & -   & -    \\
    \cline{2-12}
    \multirow{2}{*}{PHP} &   & -         & -       & -            & 42.5$_{8}$    & 50.36$_{8}$               
                             & -         & -       & -            & -    & 34.12$_{8}$     \\
                   & +PHP    & -         & -       & -            & -    & 53.9$_{8}$               
                             & -         & -       & -            & -    & 36.5$_{8}$     \\
    \cline{2-12}                                                                                  
    \multirow{2}{*}{BoT}      &  & 42.5  & -        & 47.7     & 48.93$_{8}$       & 50.4$_{8}$               
                                 & -     & -        & -        & -              & -     \\  
                         & BoT   & -     & -        & 62.5     & 66.3$_{8}$       & -               
                                 & -     & -        & -        & -     & -     \\  
    \bottomrule
    \multirow{2}{*}{CR$_{500}$} & CR    & -     & -        & -        & 54.2$_{4}$       & -               
                                        & -     & -        & -        & -                & -     \\  
                      & PHP+CR          & -     & -        & -        & 58$_{4}$         & -               
                                        & -     & -        & -        & -                & -     \\ 
    \bottomrule
    \bottomrule
    \multicolumn{12}{c}{\texttt{TheoremQA}}  \\
    \multirow{2}{*}{Source} & \multirow{2}{*}{} & \multicolumn{5}{c}{GPT-4} & \multicolumn{5}{c}{GPT-3.5-turbo}  
    \\
    \cline{3-12}
                                            &  & ZeroShot  & FewShot & ZeroShot-CoT & CoT  & Complex-CoT 
                                               & ZeroShot  & FewShot & ZeroShot-CoT & CoT  & Complex-CoT    \\
    \midrule

    \multirow{3}{*}{TheoremQA}  &  & -         & -          & -     & 43.8            & -               
                                   & -         & -          & -     & 30.2, 30.8$_{theorem}$            & -     \\

                           & PoT   & -         & 52.4       & -     & -            & -               
                                   & -         & 35.6, 35.8$_{theorem}$       & -     & -               & -      \\
    \bottomrule 
    \bottomrule
    \multicolumn{12}{c}{\texttt{MMLU}}  \\
    \multirow{1}{*}{Source} & \multirow{2}{*}{} & \multicolumn{5}{c}{GPT-4} & \multicolumn{5}{c}{GPT-3.5-turbo}\\
    \cline{3-12}
                                            &  & ZeroShot  & FewShot & ZeroShot-CoT & CoT  & Complex-CoT 
                                               & ZeroShot  & FewShot & ZeroShot-CoT & CoT  & Complex-CoT    \\
    \midrule

    \multirow{3}{*}{GPT4-report}  &  & 86.5         & 86.4          & -     & -            & -               
                                     & 70           & -             & -     & -            & -     \\

                           & BoT     & -            & -             & 90.86     & 93.42$_{5}$            & -               
                                     & -            & -             & -         & -               & -      \\
    \bottomrule

    \end{tabular}
\end{adjustbox}
\end{table*}

\section{Examples of GPT-4 with TR in \texttt{GSM8K}}

\begin{figure*}[t]
    \centering
    \includegraphics[width=\textwidth]{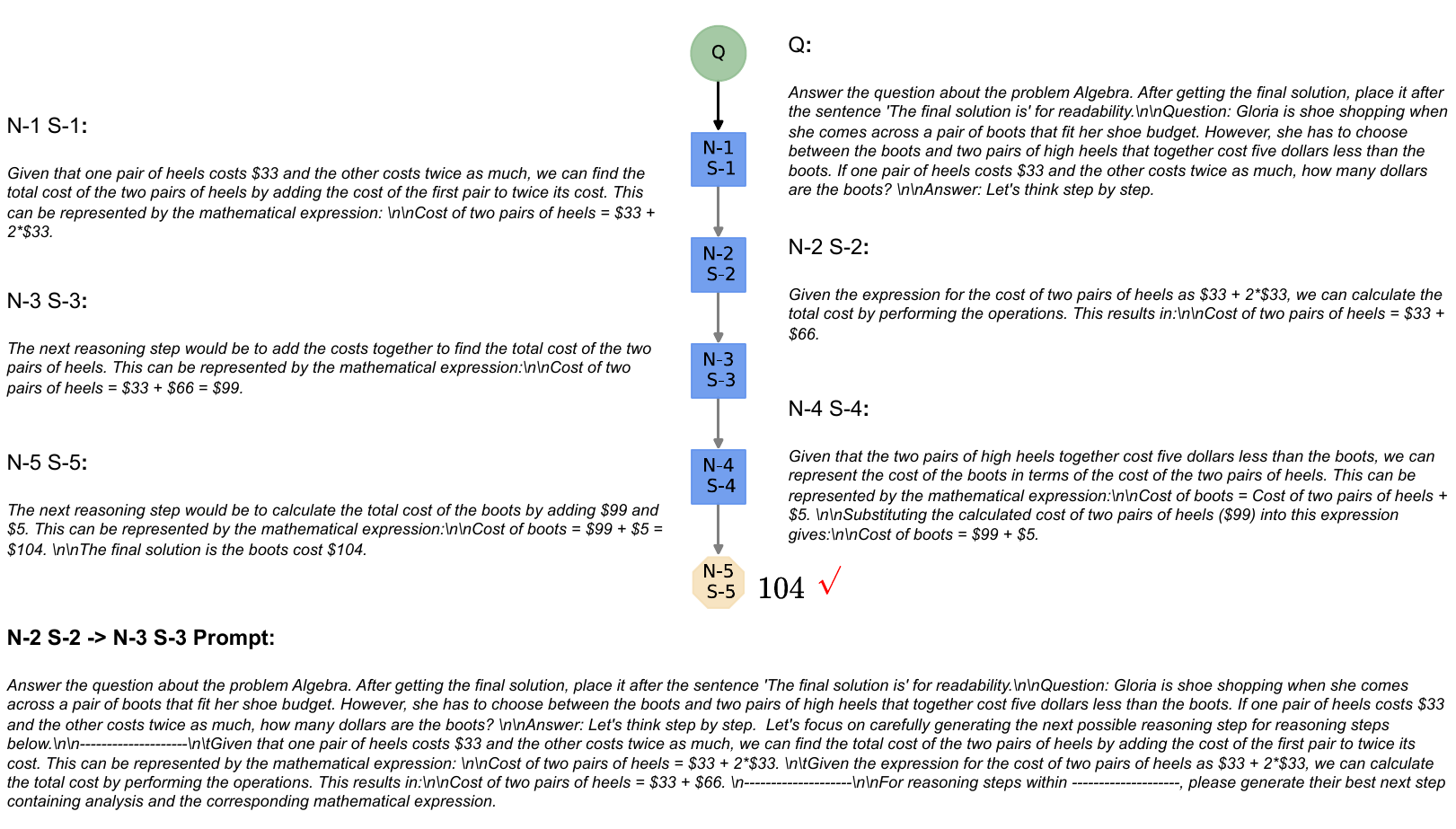}
    \caption{A simple thought structure built by GPT-4 with TR for the question from \texttt{GSM8K} dataset \cite{GSM8K-arxiv21}. This structure contains $5$ nodes, i.e. $5$ thoughts and leads to $K=1$ reasoning path towards one solution because no error is identified by the \textit{rollback controller} with GPT-4 during reasoning. }
    \label{fig:gsm8k1}
\end{figure*}

In Figure \ref{fig:gsm8k1}, we present a simple reasoning performed by GPT-4 with TR. As no bad thoughts are identified during reasoning, GPT-4 with TR directly performs correct reasoning toward a correct solution. This simple example aims to give an overview of 1). how multi-step reasoning with multi-step prompts work; 2). how to prompt LLMs to generate the next thought, such as $N-2$ $S-2$ $\rightarrow$ $N-3$ $S-3$; and 3) how LLMs with TR are able to perform normal reasoning when no rollback is triggered. Besides, as discussed in subsection \ref{subsec: overview}, LLMs with TR start from a zero-shot prompt containing only the question and task information. 

\begin{figure*}[h]
    \centering
    \includegraphics[width=\textwidth]{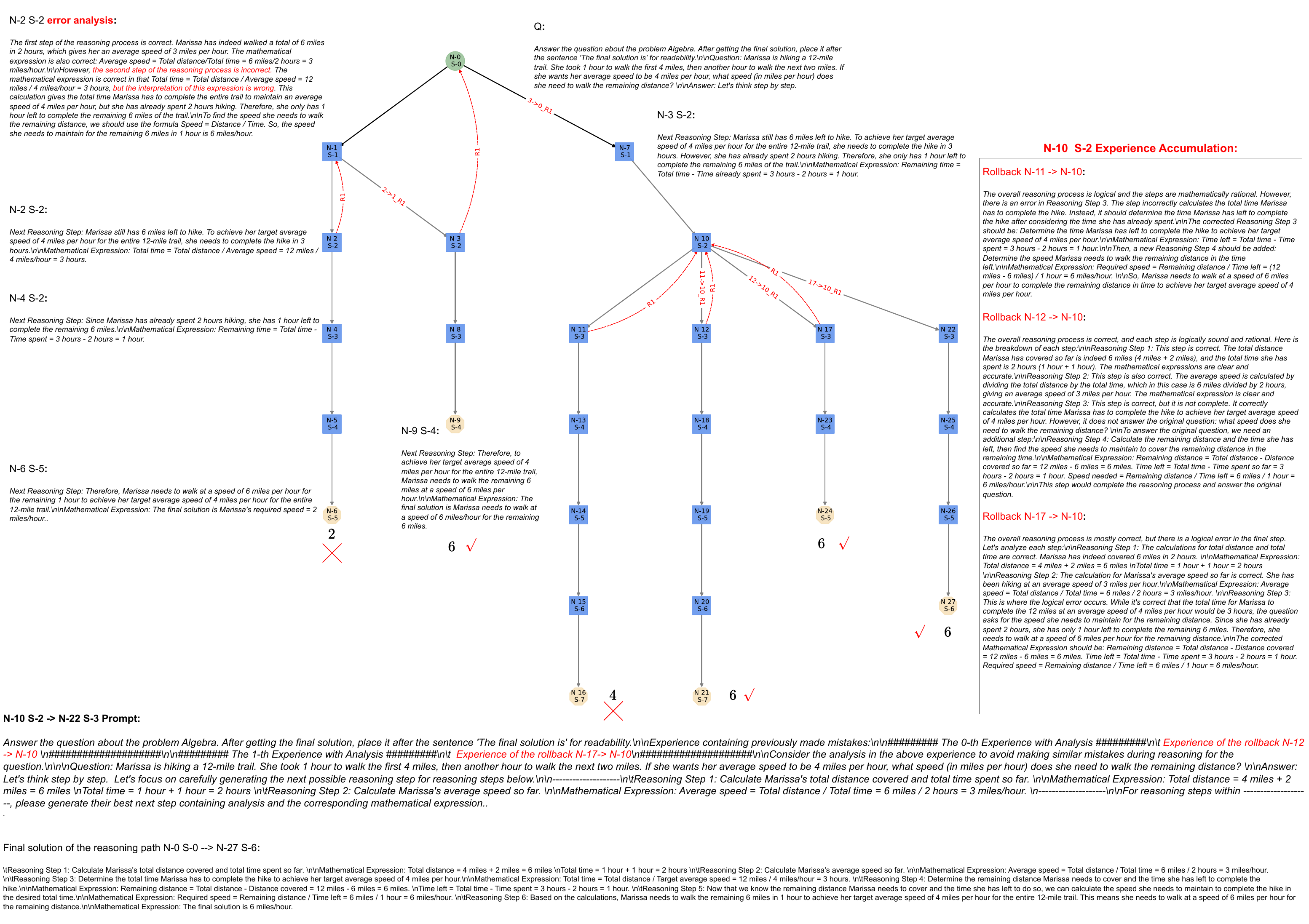}
    \caption{A slightly complex thought structure built by GPT-4 with TR for the question from \texttt{GSM8K} dataset \cite{GSM8K-arxiv21}. This structure contains $27$ nodes, i.e. $27$ thoughts and leads to $K=8$ reasoning paths towards $8$ solutions as $5$ number of rollbacks are triggered by the \textit{rollback controller} with GPT-4 during reasoning.}
    \label{fig:gsm8k2}
\end{figure*}

Then, Figure \ref{fig:gsm8k2} shows a more complex reasoning process conducted by GPT-4 with TR. In this case, \textit{rollback controller} triggers $5$ rollbacks during reasoning, leading to $8$ different solutions. After operating the experience-guided solution ensemble, we get $6$ as the final answer, which is correct. LLMs with TR first generate one thought $N-1 S-1$. Then, \textit{rollback controller} exploits LLMs to analyze the current reasoning path $N-0$ $S-1$, $N-1$ $S-1$ and $N-2$ $S-2$ and thus identifies the error in $N-2 S-2$. This triggers a rollback $N-2$ $S-2$ $\rightarrow$ $N-1$  $S-1$, leading to a new reasoning path $N-1$ $S-1$ $\rightarrow$ $N-3$ $S-2$ in which the prompt includes the error analysis, thereby revising the thought to gain a final correct solution $N-9$ $S-4$. 

As there are $3$ \emph{incoming rollbacks} for the thought $N-10$ $S-2$, the corresponding three different error analyses are accumulated as shown by the \textcolor{red}{\textbf{N-10  S-2 Experience Accumulation:}}. \textit{prompt enhancer} include these error analyses as experiences in the prompt to guide LLMs to produce correct thoughts. For instance, \textbf{N-10 S-2 $\rightarrow$ N-17 S-3 Prompt} shows that GPT-4 generates the thought $N-17$ $S-3$ from the thought $N-10$ $S-2$ with the prompt that accumulates two experiences. 

Eventually, GPT-4 with TR adaptively builds a thought structure towards generating solutions that, most of which are correct due to the continuous thought revisions via rollback of thoughts.

\section{Examples of GPT-4 with TR in \texttt{MATH}}

Limited by space, we store the detailed experimental result and visualization files of Figure \ref{fig:math1} in the folder \textit{MATH-example-1} of the supplementary. Thus, we present the question, the thoughts of correct solutions, and the values of all solutions. Specifically, in response to our discussion in section \ref{sec:tr}, we present how GPT-4 with TR is able to identify the bad thought $N-34$ $S-6$ and thus the \textit{rollback controller} gets the $N-34$ $S-6$ \textcolor{red}{error analysis}. The triggered rollback $N-34$ $S-6$ $\rightarrow$ $N-32$ $S-4$ leads to a new reasoning path $N-32$ $S-4$  $\rightarrow$ $N-35$ $S-5$, which generates a correct thought and gets the correct answer $0$.  

We specifically utilize the example in Figure \ref{fig:math2} to show how \textbf{Experience Accumulation} works in the TR framework. For the reasoning path from $N-0$ $S-0$ to $N-17$ $S-3$, there are $4$ \emph{incoming rollbacks}, including the rollback $N-3$ $S-2$ $\rightarrow$ $N-0$ $S-0$, the rollback $N-10$ $S-3$ $\rightarrow$ $N-4$ $S-1$, and the rollback $N-15$ $S-3$ $\rightarrow$ $N-11$ $S-2$. As each rollback creates an error experience from one trial of the given question, \emph{incoming rollbacks} lead to a series of experiences, as shown by \textcolor{red}{\textbf{N-11 S-2 Experience Accumulation}:}. Therefore, to generate $N-17$ $S-3$ from $N-11$ $S-2$, the \textit{prompt enhancer} includes these error analyses as experiences in the prompt, as shown in \textbf{N-11 S-2 -> N-17 S-3 Prompt:}. By learning from these experiences, LLMs are able to generate the correct thought $N-17$ $S-3$ and thus the correct answer $47$. The detailed files of this example is presented in the folder \textit{MATH-example-2} of the supplementary.

\begin{figure*}[ht]
    \centering
    \includegraphics[width=\textwidth]{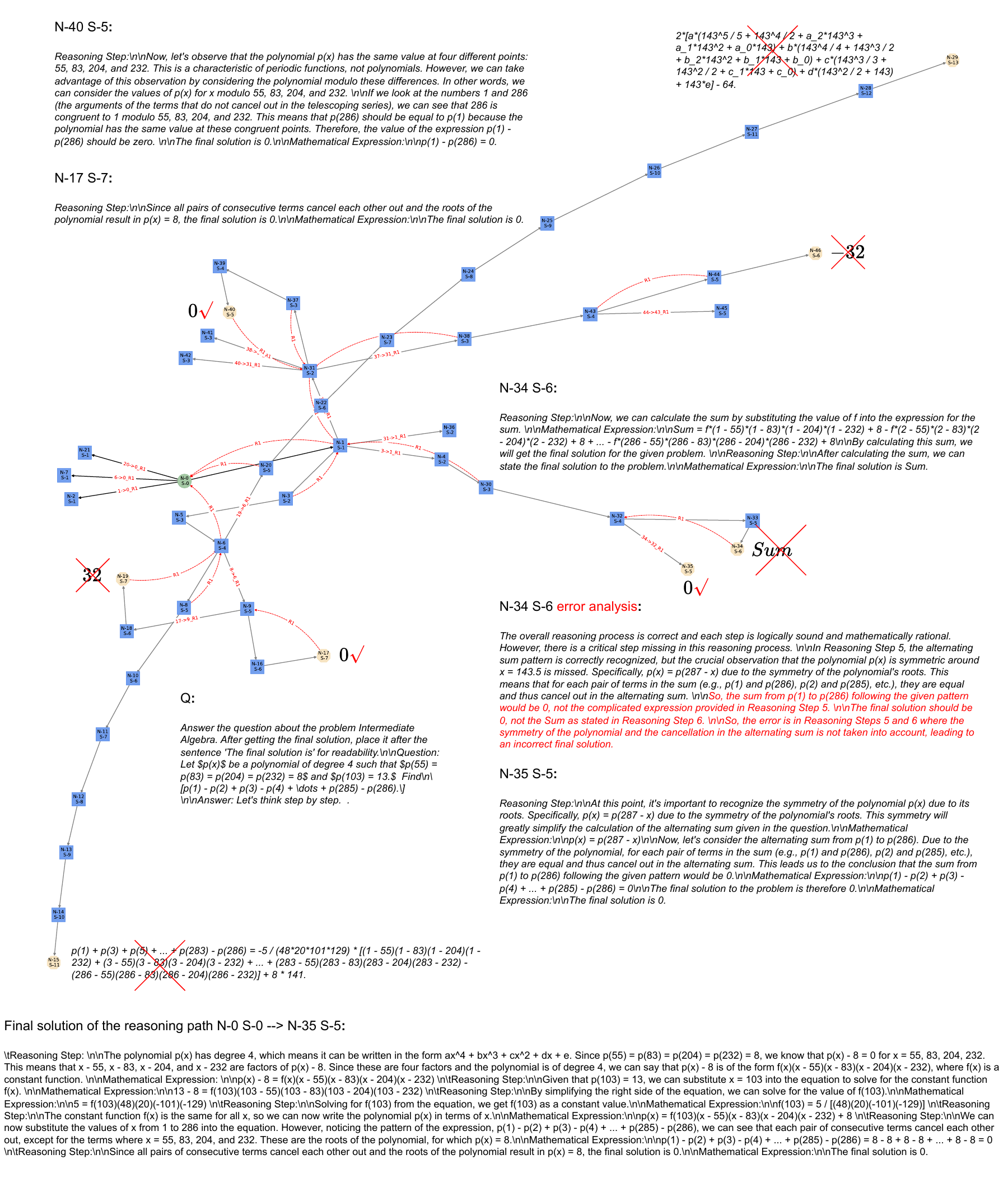}
    \caption{A complex thought structure built by GPT-4 with TR for the question from \texttt{MATH} dataset \cite{MATH-arxiv21}. This structure contains $46$ nodes, i.e. $46$ thoughts and leads to $K=8$ reasoning path towards $8$ solutions as $13$ number of rollbacks are triggered by the \textit{rollback controller} with GPT-4 during reasoning. It is plotted by based on the NetworkX lib \cite{netx-08} under the ``fdp'' layout.}
    \label{fig:math1}
\end{figure*}

\begin{figure*}[ht]
    \centering
    \includegraphics[width=\textwidth]{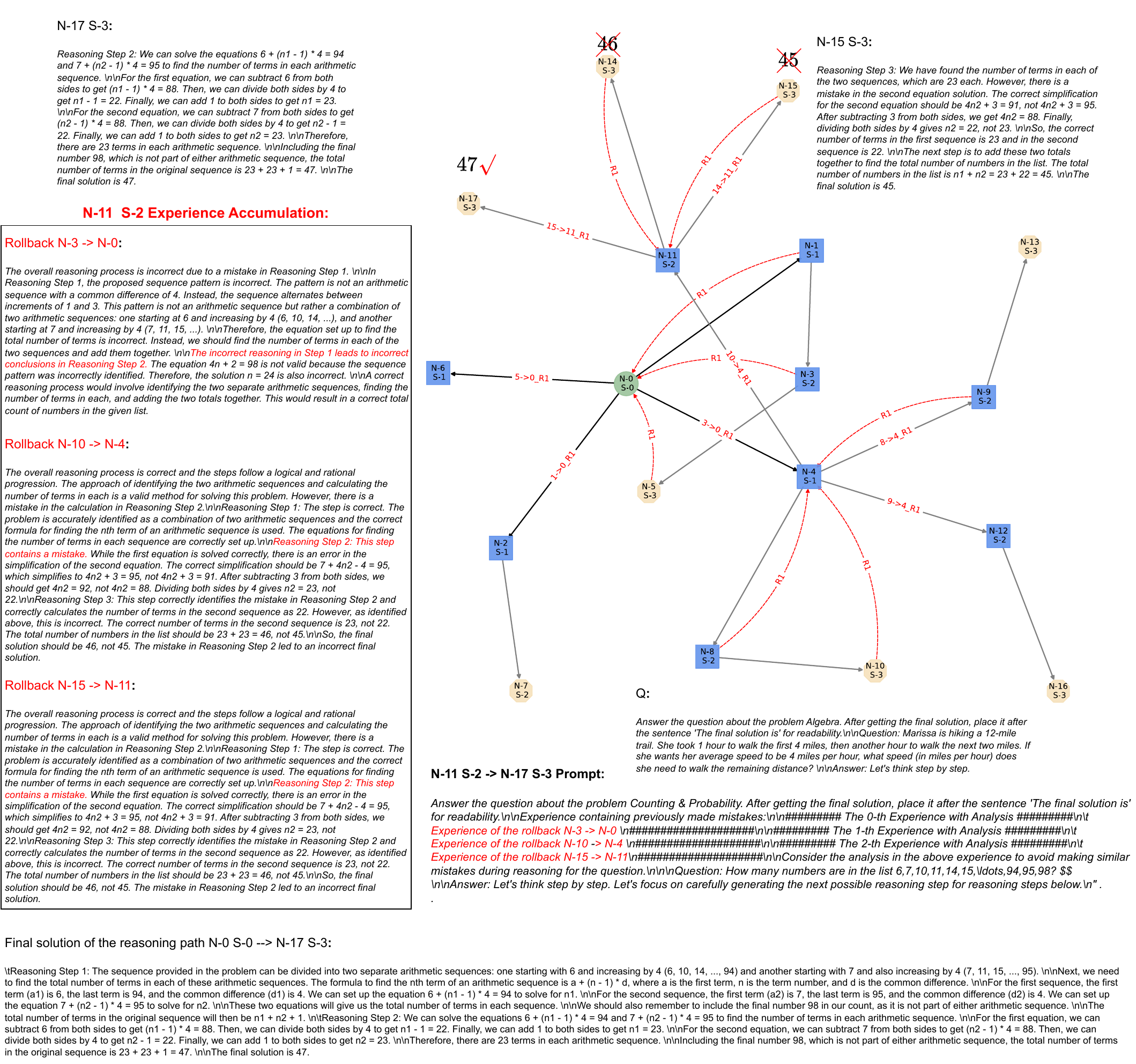}
    \caption{A complex thought structure built by GPT-4 with TR for the question from \texttt{MATH} dataset \cite{MATH-arxiv21}. This structure contains $17$ nodes, i.e. $17$ thoughts and leads to $K=8$ reasoning path towards $8$ solutions as $8$ number of rollbacks are triggered by the \textit{rollback controller} with GPT-4 during reasoning. It is plotted by based on the NetworkX lib \cite{netx-08} under the ``fdp'' layout.}
    \label{fig:math2}
\end{figure*}

\section{Examples of GPT-4 with TR in \texttt{TheoremQA}}

\begin{figure*}[ht]
    \centering
    \includegraphics[width=\textwidth]{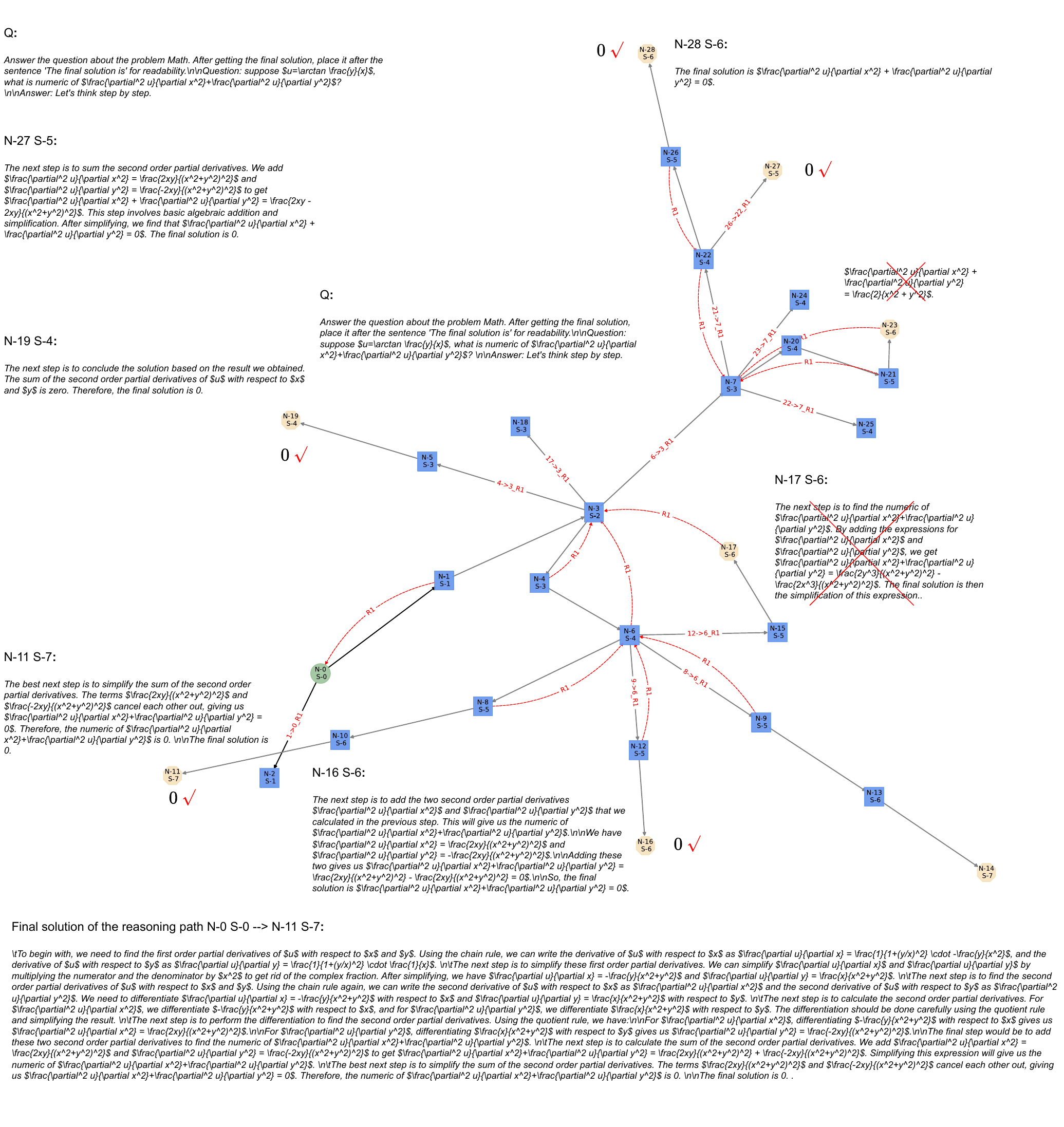}
    \caption{A complex thought structure built by GPT-4 with TR for the question from \texttt{TheoremQA} dataset \cite{TheoremQA-emnlp23}. This structure contains $28$ nodes, i.e. $28$ thoughts and leads to $K=8$ reasoning paths towards $8$ solutions as $11$ number of rollbacks are triggered by the \textit{rollback controller} with GPT-4 during reasoning.}
    \label{fig:theoremqa1}
\end{figure*}

GPT-4 with TR tends to build complex thought structures when reasoning with the challenging \texttt{TheoremQA} dataset \cite{TheoremQA-emnlp23}. As seen in Figure \ref{fig:theoremqa1}, the overall thought scale and the complexity of reasoning paths increase a lot compared to other examples. 

\begin{figure*}[ht]
    \centering
    \includegraphics[width=\textwidth]{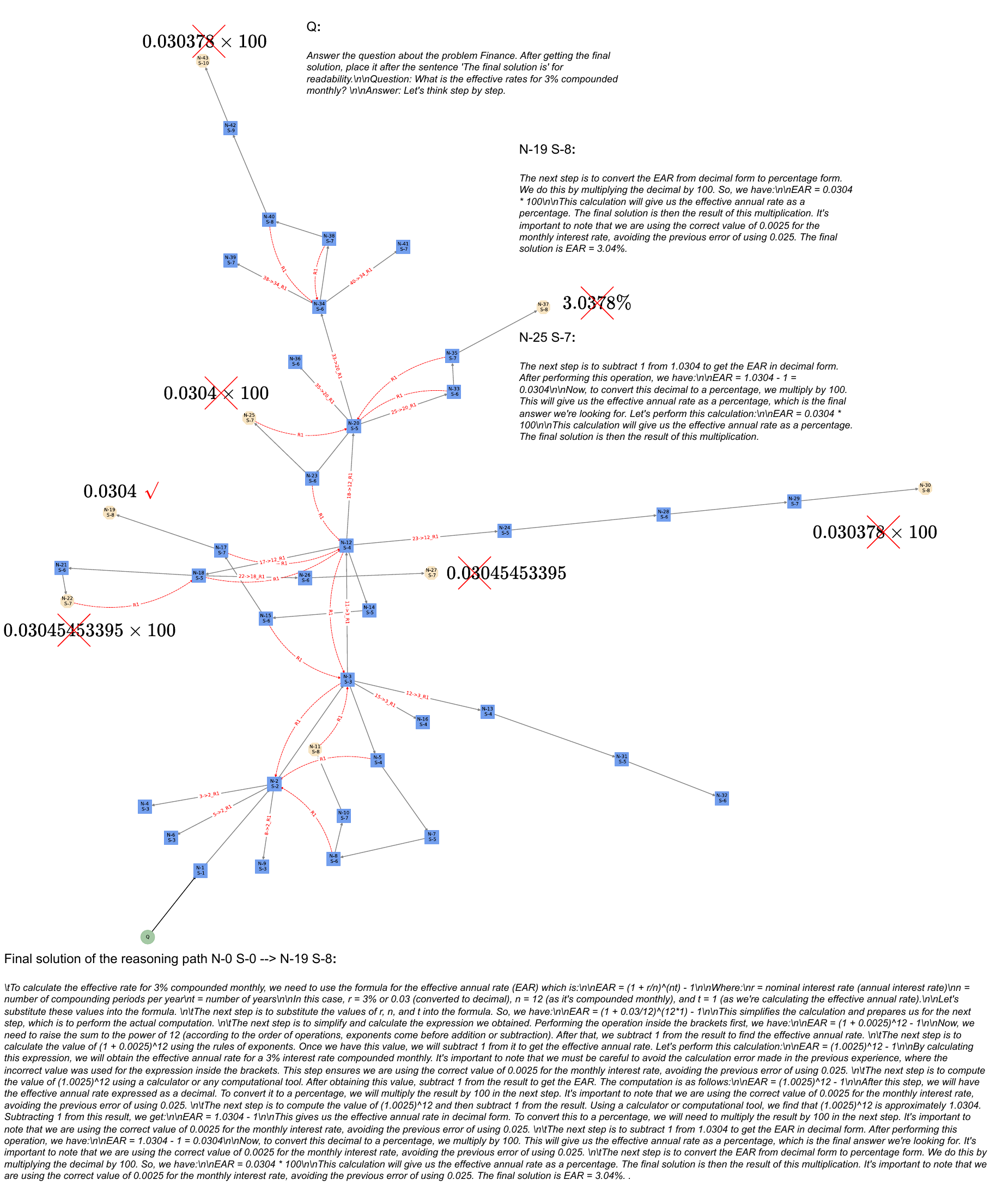}
    \caption{A complex thought structure built by GPT-4 with TR for the question from \texttt{TheoremQA} dataset \cite{TheoremQA-emnlp23}. This structure contains $43$ nodes, i.e. $43$ thoughts and leads to $K=8$ reasoning paths towards $8$ solutions as $15$ number of rollbacks are triggered by the \textit{rollback controller} with GPT-4 during reasoning.}
    \label{fig:theoremqa2}
\end{figure*}

Also, we show in Figure \ref{fig:theoremqa2} that GPT-4 with TR can build even more complex thought structures.

In these two figures, we present the obtained reasoning path toward the correct answer. For example, the ``Final solution of the reasoning path N-0 S-0 $\rightarrow$ N-11 S-7:'' in Figure \ref{fig:theoremqa1} and the ``Final solution of the reasoning path N-0 S-0 $\rightarrow$ N-12 S-7:'' in Figure \ref{fig:theoremqa2} are the correct solutions obtained by GPT-4 with TR.

\end{document}